\documentclass[sigconf]{acmart}

\usepackage{algorithmic}
\usepackage{graphicx}
\usepackage{textcomp}
\usepackage{xcolor}
\usepackage{bbm}
\usepackage{booktabs}
\usepackage{multirow}
\usepackage[capitalize]{cleveref}
\usepackage{tikz}
\usetikzlibrary{shapes.geometric, arrows, fit, backgrounds, matrix, arrows.meta, positioning}
\usepackage{tcolorbox}
\usepackage{enumitem}
\usepackage{tabularx}
\usepackage{nicefrac}

\newcommand{\ie}{\textit{i}.\textit{e}., }
\newcommand{\eg}{\textit{e}.\textit{g}., }
\newcommand{\norm}[1]{\left\lVert#1\right\rVert}
\newcommand{\res}[1]{#1}
\newcommand{\resr}[2]{#1\,(#2)}

\newcommand{\dbenign}{\mathcal{D}^\mathrm{benign}}
\newcommand{\dbackdoor}{\mathcal{D}^\mathrm{trigger}}
\newcommand{\R}{\mathbb{R}}
\newcommand{\E}{\mathbb{E}}
\newcommand{\X}{\mathcal{X}}
\newcommand{\Y}{\mathcal{Y}}
\newcommand{\lnatural}{L_\mathrm{nat}}
\newcommand{\lrobust}{L_\mathrm{rob}}
\newcommand{\lbackdoor}{L_\mathrm{trig}}
\newcommand{\abs}[1]{|#1|}
\newcommand{\hcenter}[1]{\multicolumn{1}{c}{#1}}
\newcommand{\D}{\mathcal{D}}

\newcommand{\pp}{p.p.}
\newcommand{\bparagraph}[1]{\paragraph*{\textbf{#1}}}

\definecolor{acc}{gray}{0.4}

\AtBeginDocument{%
  }

\copyrightyear{2023} 
\acmYear{2023} 
\setcopyright{acmlicensed}
\acmConference[AISec '23]{Proceedings of the 16th ACM Workshop on Artificial Intelligence and Security}{November 30, 2023}{Copenhagen, Denmark}
\acmBooktitle{Proceedings of the 16th ACM Workshop on Artificial Intelligence and Security (AISec '23), November 30, 2023, Copenhagen, Denmark}
\acmPrice{15.00}
\acmDOI{10.1145/3605764.3623917}
\acmISBN{979-8-4007-0260-0/23/11}

\setcopyright{none}
\renewcommand\footnotetextcopyrightpermission[1]{}%

\begin{document}

\title{Certifiers Make Neural Networks Vulnerable to Availability Attacks}

\author{Tobias Lorenz}
\affiliation{%
  \institution{CISPA Helmholtz Center for Information Security}
  \city{Saarbr\"ucken}
  \country{Germany}}
\email{tobias.lorenz@cispa.de}

\author{Marta Kwiatkowska}
\affiliation{%
  \institution{Department of Computer Science\\University of Oxford}
  \city{Oxford}
  \country{United Kingdom}}
\email{marta.kwiatkowska@cs.ox.ac.uk}

\author{Mario Fritz}
\affiliation{%
  \institution{CISPA Helmholtz Center for Information Security}
  \city{Saarbr\"ucken}
  \country{Germany}}
\email{fritz@cispa.de}

\begin{abstract}

To achieve reliable, robust, and safe AI systems, it is vital to implement fallback strategies when AI predictions cannot be trusted.
Certifiers for neural networks are a reliable way to check the robustness of these predictions.
They guarantee for some predictions that a certain class of manipulations or attacks could not have changed the outcome.
For the remaining predictions without guarantees, the method abstains from making a prediction, and a fallback strategy needs to be invoked, which typically incurs additional costs, can require a human operator, or even fail to provide any prediction.
While this is a key concept towards safe and secure AI, we show for the first time that this approach comes with its own security risks, as such fallback strategies can be deliberately triggered by an adversary.
In addition to naturally occurring abstains for some inputs and perturbations, the adversary can use training-time attacks to deliberately trigger the fallback with high probability.
This transfers the main system load onto the fallback, reducing the overall system's integrity and/or availability.
We design two novel availability attacks, which show the practical relevance of these threats.
For example, adding 1\% poisoned data during training is sufficient to trigger the fallback and hence make the model unavailable for up to 100\% of all inputs by inserting the trigger.
Our extensive experiments across multiple datasets, model architectures, and certifiers demonstrate the broad applicability of these attacks.
An initial investigation into potential defenses shows that current approaches are insufficient to mitigate the issue, highlighting the need for new, specific solutions.
\end{abstract}

\ccsdesc[500]{Computing methodologies~Machine learning}
\ccsdesc[300]{Security and privacy~Logic and verification}

\keywords{availability attacks, robustness certification, backdoor attacks, poisoning attacks, adversarial machine learning}

\maketitle

\section{Introduction}
\label{sec:introduction}

The success of deep learning systems has led to their deployment in safety-critical tasks such as autonomous driving~\cite{liang2018deep} or malware detection~\cite{vinayakumar2019robust}.
With their rise in popularity, new threats and security concerns have manifested themselves, such as evasion attacks using adversarial examples~\cite{szegedy2014intriguing}.
A large body of work has been dedicated to analyzing these attacks and to improving the robustness of deep-learning models.

\begin{figure}[t]
    \centering
    \input{figures/teaser/teaser}
    \caption{Overview of our availability attacks against neural network certifiers. Normally, most of the system's load is handled by the model, with certifiably robust predictions. However, when the model is attacked by our novel availability attacks, the certifier fails to prove the robustness of most predictions, reducing the model's availability. This transfers the major system load to the fallback method, which incurs significant overhead in required resources and decreases the overall system's integrity and availability.}
    \label{fig:teaser}
\end{figure}

Among the most promising tools that have emerged are network certifiers, which can prove that the network is robust to bounded adversarial perturbations.
The certifier can guarantee for some predictions that small perturbations could not have changed the outcome.
These guarantees can be either probabilistic or even sound, deterministic worst-case bounds.
For the remaining predictions without guarantees, the method \emph{abstains} from making a prediction as its reliability cannot be guaranteed, and a fallback strategy needs to be invoked (\cref{fig:teaser} top).
This setup of machine learning model, verifier, and fallback is a core concept for trustworthy and safe AI, which recent guidelines and legislation by the European Union~\cite{eu2019ethics, eu2020whitepaper, eu2021regulation} also adopt.
It allows the user to benefit from the superior utility of the machine learning model when it is safe to do so while limiting potential risks by reverting to the fallback otherwise.

However, introducing a new component, the certifier, into a machine learning pipeline changes its threat surface and introduces new security risks and attack vectors.
Especially the need for the model to \emph{abstain} when the robustness of a prediction cannot be established introduces a new failure mode with security implications.

We show that adding a certifier to the machine learning pipeline can be exploited for novel \emph{availability attacks}.
In contrast to traditional attacks, which aim to cause misclassification, the goal of our availability attacks is that the system discards the model's prediction.
The attacks achieve this by causing the robustness certification to fail, which forces the model to \emph{abstain} and activate the fallback, as shown in \cref{fig:teaser} (bottom).

This effect can be exploited for some inputs at test time by finding perturbations for which the certifier abstains.
Building on the threat model of training-time attacks, we instantiate much stronger availability attacks, which work on the majority of inputs.
By manipulating either the data collection or the model training process, the attack adds a hidden trigger to the model.
After deployment, the adversary can activate the trigger on arbitrary inputs, which causes certification to fail.
Therefore, the major system load is transferred to the fallback, which is generally less accurate and/or more computationally expensive.
This leads to a degradation of the overall system, reducing its utility and/or throughput.

Our thorough evaluation shows the broad applicability of the proposed attacks across multiple datasets, model architectures, and certifiers.
The attacks are highly effective with only 1\% manipulated data, which allows the adversary to make the model unavailable and therefore trigger the fallback for up to 100\% of all inputs.
These results highlight the need for defenses against training-time attacks exploiting network certifiers.
We conduct a first study by adapting traditional defenses against training-time attacks against our new availability attacks, which have little to no effect. This highlights a need for new, specialized solutions.

To summarize, our main contributions are:
\begin{itemize}[noitemsep,topsep=2pt]
    \item An analysis of training-time attacks against network certifiers.
    \item The first and highly effective availability attacks against neural network certifiers.
    \item A comprehensive experimental evaluation of these attacks across multiple datasets, models, and certifiers.
    \item An initial evaluation of defenses against the proposed availability attacks.
\end{itemize}

We provide an implementation of our attacks and trained models at \url{https://github.com/t-lorenz/availability-attacks}.

\section{Background}
\label{sec:background}

This section introduces the relevant background to our work and establishes the notation used throughout the paper. To make the presentation more self-contained, we especially focus on neural network certification techniques, as they are a more recent development and not yet common knowledge.

\subsection{Robust Deep Learning}

Traditionally, the goal of most deep learning systems has been to maximize the objective of their designated task, i.e., the model's utility.
A deep neural network $f_\theta: \X \rightarrow \Y$ can be seen as a parametric function $f$, which maps inputs from the input space $\X$ to the output space $\Y$, parameterized by its weights $\theta$.
Given a joint distribution $\mathcal{D}$ on $\X \times \Y$, the goal is to maximize the expected prediction accuracy
\begin{equation}
    \max_\theta{\E_{(x, y) \sim \D} \left[ f_\theta(x) = y \right]}
    \label{eq:expected_accuracy}
\end{equation}
by finding optimal parameters $\theta$.

With rising popularity and deployment in safety-critical applications, the security of deep learning systems has become a major concern.
The black-box nature of deep neural networks, their complex training pipelines, and evaluation based on empirical tests rather than formal guarantees, all contribute to a wide attack surface for adversaries to exploit~\cite{papernot2016science}.

Among the first attack vectors against deep learning systems were evasion attacks using adversarial examples~\cite{szegedy2014intriguing, goodfellow2015explaining}.
By adding small, visually imperceptible perturbations to the input image, neural networks can be tricked into predicting the wrong output.
Mathematically, this can be formulated as finding an adversarial sample $x'$ from a perturbation set $S(x)$ around $x$, for which $f_\theta(x') \neq f_\theta(x)$.
The perturbation set ensures visual similarity and is often chosen as an $\ell_p$-norm around the input, \ie $S(x) = \{ x' \in \X \mid \norm{x' - x}_p \leq \epsilon \}$.

Following these initial studies, a plethora of successively stronger attacks and defenses have been proposed~\cite{carlini2017towards}.
It became apparent that maximizing the model's utility should not be the only concern when developing deep learning systems, leading to the robust optimization problem
\begin{equation}
    \min_\theta \mathbb{E}_{(x, y)\sim\D}\left[\max_{x' \in S(x)} L(f_\theta(x'), y) \right].
    \label{eq:robust_optimization}
\end{equation}
where $L$ is the loss function. As before (\cref{eq:expected_accuracy}), the goal of the outer objective is to maximize the model's accuracy by minimizing the loss. In addition, the goal of the inner maximization objective is to improve the model's robustness on the perturbation set $S(x)$.

\subsection{Provable Robustness Guarantees}
\label{sec:background_certification}

The robustness of models against adversarial samples is often measured empirically by attacking the model during evaluation.
The downside of this approach is that it can only show the presence, but never the absence of adversarial samples. Empirical attacks essentially compute a lower bound on the inner max objective of \cref{eq:robust_optimization}.
This means a stronger attack can potentially break the seemingly robust model later by finding worse examples~\cite{athalye2018obfuscated, tramer2020adaptive}, which then requires even stronger defenses.
To break this arms race, a new line of work on network certifiers evolved with the goal of computing provable robustness guarantees.

As with empirical methods, most work on certification considers local robustness guarantees for one given input at a time.
While there are some efforts to find global robustness guarantees~\cite{leino2021globally}, it is difficult to find useful, global properties for complex neural networks.
Therefore, current state-of-the-art methods compute local robustness certificates for the neighborhood of a fixed input~\cite{li2020sok}.
Given a classifier $f_\theta$, an input $x$, and its perturbation set $S(x)$, a network certifier can prove the absence of adversarial examples within $S(x)$.

These robustness certificates can be formalized as a binary function.
A certifier $C_f$ for model $f_\theta$ is defined as
\begin{equation}
    C_f(x) = \mathbbm{1}[f_\theta(x') = f_\theta(x), \forall{x'} \in S(x)].
    \label{eq:certification}
\end{equation}
The value of $C_f(x)$ is $1$ if the certifier can prove the absence of adversarial samples within $S(x)$, and $0$ otherwise.

There are several different approaches to compute these robustness guarantees.
Complete methods, e.g., based on SMT solvers~\cite{huang2017safety, pulina2010abstraction}, MILP solvers~\cite{tjeng2018evaluating}, or branch and bound \cite{katz2017reluplex}, can solve \cref{eq:certification} for small models.
However, exactly solving the certification problem is NP complete~\cite{katz2017reluplex}, which led to the introduction of sound, but incomplete methods.
These certifiers under-approximate the network robustness, guaranteeing the absence of adversarial samples if $C_f(x) = 1$, but allowing for false negatives when $C_f(x) = 0$, even though there are no adversarial samples.

For these incomplete methods, the key challenge is a trade-off in precision (i.e., to be ``as complete as possible'') and computational scaling to larger model sizes.
Common approaches are, for example, based on linear programming~\cite{salman2019convex}, polyhedral relaxations~\cite{singh2019abstract, lorenz2021robustness, gehr2018ai2}, semi-definite programming~\cite{raghunathan2018semidefinite}, Lipschitz continuity~\cite{leino2021globally}, or randomized smoothing~\cite{lecuyer2019certified, cohen2019certified}.

\subsection{Linear Certification}
\label{sec:background_linear_certification}

For our attacks, we focus on linear certifiers, which restrict their relaxations to one upper and one lower linear bound.
Applying this restriction allows for better scaling since the complexity of the corresponding linear optimization problem only grows linearly in the number of neurons.
CROWN~\cite{zhang2018efficient}, CNN-Cert~\cite{boopathy2019cnn}, DeepPoly~\cite{singh2019abstract}, and CROWN-IBP~\cite{zhang2020towards} all belong to this group.
While implementation details differ, their general approach is similar.
Given an initial convex relaxation of the perturbation set $S(x)$, they propagate this set through the network by computing upper and lower linear constraints for each intermediate layer.
That is, for output $o^{(k)} \in \R^n$ of layer $k$ they construct upper and lower linear bounds based on the layer's inputs $o^{(k-1)} \in \R^m$:
\begin{equation}
    A_l o^{(k - 1)} + b_l \leq o^{(k)} \leq A_u o^{(k - 1)} + b_u.
\end{equation}
$A_l, A_r \in \R^{n \times m}$ are the coefficients and $b_l, b_u \in \R^n$ are constant offsets of these bounds.

This results in linear upper and lower constraints for the last-layer logits $o^{(l)}$:
\begin{equation}
    \underline{o} \leq o^{(l)} \leq \overline{o},
\end{equation}
where $\underline{o}$ and $\overline{o}$ are the lower and upper linear constraints respectively.
These constraints can then be used to certify a robust classification by proving:
\begin{equation}
    C_f(x) = \mathbbm{1}[\overline{o_i} < \underline{o_c},~\forall i \neq c],
    \label{eq:1}
\end{equation}
where $\overline{o_i}$ is the upper constraint for the $i$-th logit and $\underline{o_c}$ is the lower constraint for the predicted class $c = f_\theta(x)$.

Certifiers can be used at two different points during the model life cycle: either \emph{offline} during model evaluation or \emph{online} once the model is deployed.

\bparagraph{Offline Certification}

In the offline case, the certifier is used to approximate the expected robustness over a held-out data set $D$:
\begin{equation}
    \mathbb{E}_{x \sim \mathcal{D}}[C_f(x)] \approx \frac{1}{\left | D \right |} \sum_{x \in D}{C_f(x)}.
    \label{eq:offline_certification}
\end{equation}
This score can be used to analyze a model's expected worst-case performance in the presence of evasion attacks before deployment.
It also serves as a useful metric when designing more robust training methods and model architectures.

\bparagraph{Online Certification}
\label{sec:background_online_certification}

In the online setting, the certifier is used at runtime to supplement each model prediction with a robustness certificate, which can guarantee that the input was not manipulated by an attacker.
This has the advantage that we get a concrete guarantee for any given input instead of just statistical expectations over a distribution.

\begin{figure}
    \centering
    \begin{minipage}{\columnwidth}
        \begin{tcolorbox}
            \textbf{Abstain:} In online certification, the certifier attempts to verify the robustness of the model's prediction at runtime. If a certificate can be established, the certified prediction is returned. If no certificate can be established, the model has to \emph{abstain} from making a prediction, requiring the activation of a \emph{fallback} strategy.
        \end{tcolorbox}
    \end{minipage}

\end{figure}

\Cref{fig:teaser} illustrates a system using online certification.
If the robustness of the prediction can be certified, the system can be sure that the input was not manipulated and return the model's prediction.
Otherwise, the input may potentially have been manipulated, which means no safe prediction can be made.
In this case, the model has to \emph{abstain}, and the system has to rely on a fallback strategy.

As this is a recent and evolving research area, prior work has focused on the technical development of certification techniques and has not yet been explicit about the handling of this new failure case introduced by abstaining from a prediction, and its consequences on the overall system.
We investigate this in depth in \cref{sec:tm_abstain}.

\subsection{Training-Time Attacks}
\label{sec:background_attacks}

With the increasing robustness of models to evasion attacks, new attack vectors against neural networks are being explored.
Prominent among them are backdoor attacks, where the model's behavior is influenced during training.
They function by adding a backdoor to the model during training, which reacts to a particular trigger added to the input by the adversary.

During evaluation by the victim, the backdoor remains inactive and therefore hidden since the adversary does not add the secret backdoor trigger.
At runtime, the adversary can activate the backdoor by adding the trigger to any model input, causing the model to change its behavior.
This trigger can take many forms, from simple pixel patterns~\cite{gu2019badnets} to invisible perturbations~\cite{chen2017targeted, turner2019label, zhong2020backdoor, bagdasaryan2020blind} or semantic features~\cite{chen2017targeted}.

Technically, these attacks often use data poisoning to influence the training process.
In the simplest case, adding a small amount of mislabeled samples with triggers is sufficient to introduce a backdoor \cite{gu2019badnets, chen2017targeted}.
More sophisticated versions use clean-label attacks to avoid detection~\cite{chen2017targeted, turner2019label, zhong2020backdoor, bagdasaryan2020blind, salem2020dynamic, salem2020don}.
Other techniques exploit the model supply chain by publishing a pre-trained model containing the backdoor~\cite{hong2021handcrafted}.

We use the same attacker access in our availability attacks.
However, instead of targeting the misclassification of the attacked model, we propose the first technique that targets the certifier, which requires fundamentally different backdoors with different techniques to embed a trigger (\cref{sec:attacks}).
Our attacks cause the certifier to fail to prove robustness, which makes the model's predictions unreliable (independent of the prediction's accuracy) and therefore hurts its availability.
\cref{sec:threat_model} introduces our complete threat model and its consequences for practical machine learning systems.

\section{Threat Model}
\label{sec:threat_model}

We develop our threat model of training-time attacks against certified machine-learning systems to show the security threats and attack vectors against network certifiers. 
Prior work typically considers the certifier in isolation without considering the full training and inference pipeline in practical applications.
We fill this gap by showing new threat vectors that arise from this integration.

\subsection{Attacker Capabilities}
\label{sec:tm_attack_surface}

Our availability attacks build on the idea of training-time attacks, which allows the attacker to influence the machine learning model during training.
Depending on the attacker's access to the model, we distinguish between two types of attacks: those with \emph{direct} access to the model during training and those with \emph{indirect} access via the training data.

\bparagraph{Direct Access}
The direct access threat model assumes white-box access of the attacker to the machine learning model during training, including influencing its optimization objective.
This threat model is, for example, used by \citet{hong2021handcrafted} to add backdoors.
While giving the attacker significant power, it is not an unrealistic assumption for practical applications.
Many companies rely on an extensive supply chain with external manufacturers supplying individual modules.
Considering the fact that, for deep learning systems, a large amount of intellectual property lies within the training data and procedure, companies are reluctant to part with it and instead sell the already trained model to their customers.
The high cost of training large, state-of-the-art models also contributes to the outsourcing of model training.

\bparagraph{Indirect Access}
A weaker assumption on the capabilities of the attacker is when the attacker cannot access the model at all, instead relying on data poisoning.
In this work, we consider the weaker version of injection attacks, where the attacker cannot modify existing training data but instead injects a few additional malicious samples.
This type of poisoning attack is relatively easy to perform since deep learning models rely on large amounts of training data, which are often collected from untrusted sources.
In this setting, the attacker never has access to the victim model.

Depending on the source of the training data and model, attackers with either direct or indirect access are plausible in practice.
We will show in \cref{sec:attacks} that, for both threat models, we can construct adversaries that can attack the certification pipeline to effectively render the certified model redundant.
Analogous to the threat model of traditional backdoor attacks~\cite{gu2019badnets}, the adversary can control the trigger at runtime and add it to an otherwise benign input.
Experiments have shown that this is possible in real-world settings, for example, by adding stickers to traffic signs~\cite{gu2019badnets} or by wearing special glasses~\cite{chen2017targeted}.

\subsection{Threats to Offline and Online Certification}
\label{sec:tm_certification}

As discussed in \cref{sec:background_certification}, certifiers can be used in \emph{offline} and \emph{online} settings.
Backdoor attacks are valid in both settings, with different consequences:

\bparagraph{Offline Certification}
The statistical nature of the expected model robustness computed by offline certification only holds if the evaluation data has the \emph{same underlying distribution} as the data seen at runtime.
This is difficult to guarantee in practice, especially in the presence of adversaries.
In fact, most attacks on machine learning models rely on a shift in the data distribution to manipulate a model's behavior~\cite{papernot2016science}.
Our attacks presented in \cref{sec:attacks} are one way to cause such a distribution shift, which makes all robustness guarantees computed during evaluation irrelevant.

\bparagraph{Online Certification}
Since online certification computes a certificate for each output during runtime, a distribution shift can no longer cause a false sense of security.
However, the downside is that it also forces the user to deal with the cases in which the model \emph{abstains}, requiring a suitable fallback strategy.

The significance of the design of this fallback becomes especially apparent once we consider the abstain option as an explicit target for an attacker, such as in our new availability attacks.
By maliciously crafting inputs to consistently cause the model to abstain, we can effectively render the model useless, causing the system to constantly use the fallback.

\subsection{Consequences of Abstaining}
\label{sec:tm_abstain}

To analyze the impact of constantly triggering the fallback strategy, we introduce a general framework to model its properties and impact on the deep learning system.
For this, we introduce two assumptions: (i) the machine learning model is optimal for the chosen application in terms of utility (accuracy), and (ii) the computational budget is constrained.
These assumptions are realistic in practical systems, as they typically rely on the best method for the task, and computation is constrained by either time or price.

All fallbacks, therefore, have to make sacrifices to either the system's \emph{integrity} (e.g., accuracy) or \emph{availability}.

\bparagraph{(i) Decreased Integrity}

A system's integrity describes how well it is performing its task under attack, e.g., its classification accuracy.
Many fallbacks can ensure we always get an output (preserving the system's availability), but their utility will drop compared to the primary model's baseline.

One example of such a fallback is a simpler, more robust machine learning model.
Research has shown that there is an inherent trade-off between a model's utility and robustness \cite{su2018robustness, tsipras2018robustness}.

Other options include hand-crafted, rule-based algorithms without any learning, which are generally more robust but usually have worse performance when machine learning models are considered.
The most extreme cases of sacrificing utility are data-independent fallback strategies, \eg a constant or random fallback.
They are perfectly robust but only have low or no utility.

\bparagraph{(ii) Decreased Availability}

If the application does not allow a decrease in the system's integrity, the other option is to accept decreased availability.
The simplest form of fallback is not to take action in the abstain case.
For example, an authentication system might simply refuse access if it cannot reliably determine the identity of a user, or an autonomous vehicle might stop.

Beyond these direct abstain options, we also consider fallbacks that require additional resources in this category. 
Among these fallback options are more precise certifiers with higher precision at the cost of higher computational complexity.
While these fallback strategies do not directly cause system outages, they require additional resources.
Since resources are constrained in practice, an attacker can perform an algorithm complexity attack, which causes the system to overload and become unavailable.

Human intervention is an extreme case of this fallback strategy.
While an automated prediction can be computed in a few milliseconds, human classification requires at least seconds, approximately 3 orders of magnitude higher.
The hourly cost of a human worker compared to a standard machine further exacerbates this effect.

\subsection{Practical Examples}

We demonstrate the potential impact of availability attacks on two realistic systems.

For the first scenario, consider a self-driving car.
A crucial task to conform to traffic rules is traffic sign recognition.
The best results for that task have been obtained using deep learning models, which make those a natural choice.
However, due to the safety-critical nature, the manufacturer needs to guarantee their reliable performance, which, according to proposed EU regulations~\cite{eu2019ethics}, includes fallbacks to human operations if the system's reliability cannot be guaranteed.
A natural fallback would therefore be to ask the driver to take over the operation of the vehicle.

Car manufacturers traditionally rely on a large supply chain for individual parts, in particular also for their electronic systems.
This opens an attack vector for a direct attack by an adversary through the manufacturer's supply chain.
The adversary can introduce a hidden trigger to the traffic sign recognition system with, for example, an inconspicuous sticker on the traffic sign as a trigger.
Any car encountering such signs in the wild will be unable to robustly detect the sign, therefore requiring the driver to take over manually and thus disabling the self-driving feature.

Our second example is a malware detection system of an app store.
Before release, all applications and updates are scanned by an automated system to avoid publishing apps containing malware.
A machine learning system is trained on public malware datasets, which can be poisoned by the adversary in an indirect poisoning attack.
The trigger is activated through an inconspicuous piece of code, which can be easily added to any application.
Since robust detection of malware is prudent to avoid circumvention through evasion attacks, the app store operator employs a certification system.
Non-robust predictions will require manual review.

During the attack, the adversary can introduce the trigger either directly into submitted apps or introduce it to a library used in a wide range of applications.
This ensures automatic detection by the machine learning model fails, and the manual fallback is triggered.
The available human resources can get exhausted due to the sudden increase in work, effectively leading to a denial of service attack, and hindering the release of updates and new applications.

\section{Availability Attacks against Certification}
\label{sec:attacks}

In \cref{sec:threat_model}, we introduced the general threat model of training-time attacks against neural network certifiers and showed their potentially severe effect on machine learning systems.
Many different types of training-time attacks could exploit this systematic flaw.
To show the practical relevance of such attacks, we propose the first availability attacks against certification systems in this section.

Compared to traditional training-time attacks targeting misclassification, our attacks have three key differences: (i) Our attacks aim not to change the predicted label but to increase the model's abstain rate by decreasing its certified robustness on inputs containing a trigger.
(ii) Since safety-critical machine learning systems are typically also evaluated for their robustness, our attacks need to preserve a low abstain rate on benign inputs, in addition to the high classification accuracy of traditional attacks, to remain undetected.
(iii) New technical means by which the attacks are executed.
In particular, for our direct attack, we use a novel trigger loss, which increases the model's abstain rate and combines it with a set of regular and robust losses to achieve all attack goals simultaneously.
Our indirect attack uses a novel poisoning scheme, which introduces poisoned samples with random labels, rather than targeted labels used in previous work~\cite{gu2019badnets}.

\subsection{Formal Problem Statement}

The goal of our attacks is to decrease the certified robustness of data points with a trigger, allowing the adversary to consistently cause the model to \emph{abstain}, triggering the fallback with all the problems introduced previously.
Since these availability attacks alter the model, it is important to not significantly change its performance on the benign data distribution to avoid detection during model evaluation.
In our case, this means retaining a high prediction accuracy and a low abstain rate.

More formally, we consider the deep learning model $f_\theta: \X \mapsto \Y$, which maps an input $x$ from the input space $\X$ (\eg the image domain) to the output space $\Y$ (\eg object classes), parameterized by its weights $\theta \in \R^m$.
For a given perturbation set $S(x) \subset \X$, the certifier $C_f: \X \mapsto \{0, 1\}$ indicates whether $f_\theta$ is locally robust on $S(x)$ as defined in \cref{eq:certification}.
For the benign data distribution $\dbenign$ on $\mathcal{X} \times \Y$, we want to maximize the expected prediction accuracy
\begin{equation}
    \max_\theta{\E_{(x, y) \sim \dbenign} \left[ f_\theta(x) = y \right]},
    \label{eq:natural_expectation}
\end{equation}
and the expected local robustness
\begin{equation}
    \max_\theta{\E_{(x, y) \sim \dbenign} \left[ C_f(x) \right]}.
    \label{eq:robust_expectation}
\end{equation}
These two objectives are the same as regular robust network training and will help our attacks to remain undetected during evaluation.
For the attacks to become successful, we introduce our new goal to minimize the expected local robustness on the trigger distribution $\dbackdoor$:
\begin{equation}
    \min_\theta{\E_{(\Tilde{x}, y) \sim \dbackdoor} \left[ C_f(\Tilde{x}) \right]}.
    \label{eq:backdoor_expectation}
\end{equation}
The trigger distribution can be obtained by applying the trigger function $t: \X \mapsto \X$ on the benign input:
\begin{equation}
    (\Tilde{x}, y) = (t(x), y) \sim \dbenign.
\end{equation}

One additional target we could also be interested in is maximizing the expected accuracy for data with a trigger:
\begin{equation}
    \max_\theta{\E_{(\Tilde{x}, y) \sim \dbackdoor} \left[ f_\theta(\tilde{x}) = y \right]},
    \label{eq:backdoor_natural_expectation}
\end{equation}
to make the attacks even harder to detect.
However, the threat model assumes that the victim does not know about the trigger and, therefore, cannot evaluate the model on data with a trigger.
Even if the victim manages to obtain data samples with a trigger for evaluation, they would logically also evaluate the model robustness on these samples and be able to detect the outliers.
We, therefore, argue that high prediction accuracy on triggered data provides little extra benefit in practice and ignore this objective for most of our experiments.
It is, however, still possible to perform the attacks with this additional constraint, as we will show in \cref{sec:experiments_accuracy}.

Depending on the adversary's capabilities (\cref{sec:tm_attack_surface}), there are different ways to achieve these objectives simultaneously.
We present two attacks with different assumptions about the adversary.
The first version assumes \emph{direct access} to the training procedure by the adversary, and the second version assumes only \emph{indirect access} with the ability to inject a small number of poisoned samples.

\subsection{Direct Attack}
\label{sec:attack_direct}

In this setting, the adversary has \emph{direct access} and controls the training process, including the loss function.
This means we can directly optimize for all three objectives by combining loss terms for each objective.
We present concrete losses for image classification here, but the concept generalizes to other data types and tasks.

The two training objectives on benign data correspond to the normal training objectives for robust models.
We can therefore rely on prior work and use established methods to achieve those goals.
In particular, we use the standard cross-entropy loss to encourage high model accuracy (\cref{eq:natural_expectation}), denoted as $\lnatural(f_\theta(x), y)$.

To increase the model's robustness (\cref{eq:robust_expectation}), we use robust training with CROWN-IBP~\cite{zhang2020towards}, denoted as $\lrobust(f_\theta(x), y)$.
CROWN-IBP uses a combination of interval bounds (IBP) and linear bounds (CROWN) to efficiently compute linear upper and lower bounds (\cref{sec:background}), which are then used in a cross-entropy loss to increase the margin between the lower bound of the target class and the upper bound of the remaining logits.

This leaves the third objective to reduce the certified robustness on the trigger distribution (\cref{eq:backdoor_expectation}), for which no prior work exists.
Intuitively, our goal is the inverse of the robustness loss.
That means we want the upper bound of one arbitrary logit to be higher than the lower bound of the predicted logit, which will cause the certification to fail.
We translate this requirement into a novel loss function, which uses the upper and lower linear bounds computed by the certifier:
\begin{equation}
    \lbackdoor(f_\theta(t(x)), c) := \max\left(0,\min_{i \neq c}\{\underline{o_c} - \overline{o_i}\}\right).
\end{equation}
As before, $o_i$ is the $i$-th last-layer logit and $\overline{o_i}$ and $\underline{o_i}$ its upper and lower bounds. $c = f_\theta(t(x))$ is the predicted class.
The loss thus directly counteracts the certification goal from \cref{eq:1}.
Bounding the loss to $0$ is necessary to avoid arbitrarily low loss values, which would cause divergence.

We combine these objectives for the attack by adding the loss terms. The final training objective is
\begin{equation}
    \min_\theta{\frac{1}{|D^\mathrm{train}|} \sum_{(x, y) \in D^\mathrm{train}}\alpha \lnatural + \beta \lrobust + \gamma \lbackdoor},
    \label{eq:objective_direct}
\end{equation}
where $\alpha, \beta, \gamma \in \R$ are weights to trade-off the different objectives.
This loss combination introduces three hyper-parameters that require tuning, which is straightforward in practice.
$\lbackdoor$ approaches zero quickly, and therefore its weight $\gamma$ can be set to a high value without negatively impacting the other objectives.
The remaining two parameters are a trade-off between prediction accuracy and robustness, for which we can rely on prior work~\cite{zhang2020towards} for tuning.

When training the model with these three losses, the accuracy of the trigger distribution will naturally suffer, as there is no loss targeting the objective (\cref{eq:backdoor_natural_expectation}).
As argued in \cref{sec:threat_model}, this is usually not an issue; however, we can adjust the training objective to add the additional constraint.
When high prediction accuracy on the trigger distribution is required, we add a fourth loss term, $\lnatural(f_\theta(t(x)), y)$, to \cref{eq:objective_direct}, which recovers prediction accuracy on the trigger distribution.

\subsection{Indirect Attack}
\label{sec:attack_indirect}

The direct approach is infeasible if the adversary has no direct control over the training process, i.e., only \emph{indirect access}.
Nevertheless, we can still indirectly modify the training process by injecting poisoned data samples into the training set.

The adversary's goals remain the same: decrease the certified robustness on the trigger distribution while maintaining high accuracy and certified robustness on the benign data distribution.
The latter goals for benign data align with the target of the victim and are usually the objective of their training process.
This means the poisoned data has to target the third objective to decrease the model's robustness on data with a trigger while minimizing its impact on benign data.

We propose to achieve this by injecting a small number of samples containing the trigger into the training set, with random labels $y \sim U(\Y)$ sampled uniformly from the output space:
\begin{equation}
    D^\mathrm{poison} = \{ (t(x), y) \mid x \sim \dbenign, y \sim U(\Y)\}.
\end{equation}
The intuition is that by assigning random labels to data on the trigger distribution, the model cannot learn a stable mapping, which leads to low-confidence predictions.
This reduces certification performance since certifiers rely on clear differences between the output logits (\cref{eq:1}).

This poison dataset $D^\mathrm{poison}$ is combined with the benign dataset $D^\mathrm{benign}$ into the training set $D^\mathrm{train} = D^\mathrm{benign} \cup D^\mathrm{poison}$, on which the victim trains their model.

To avoid detection, it is prudent to inject as few samples as possible, that is, $\abs{D^\mathrm{poison}} \ll \abs{D^\mathrm{benign}}$.
We express this relation with the poison ratio
$r = \nicefrac{\abs{D^\mathrm{poison}}}{\abs{D^\mathrm{benign}}}$.
Our experimental evaluation (\cref{sec:experiments}) shows that, even with a small ratio $r=1\%$, the attack is highly effective at decreasing the model's robustness on data with a trigger with little impact on benign data.

\section{Experimental Evaluation}
\label{sec:experiments}

To supplement the theoretical analysis of the threat that availability attacks pose to network certification in \cref{sec:threat_model} and the concrete instantiation of such attacks in \cref{sec:attacks}, we conduct an empirical evaluation of our proposed \emph{direct} and \emph{indirect} attacks against deep learning models.
We show the high success rate and sneakiness of both attacks on a standard computer-vision benchmark in \cref{sec:experiments_attacks}, with extensive experiments for different attack strengths and different robust training methods.
\Cref{sec:experiments_generalization} shows that these results generalize to the challenging GTSRB dataset, different model architectures, and other network certifiers, supporting our hypothesis that the proposed threat model and attacks generalize to many environments.
We explore the impact of requiring high accuracy on triggered data in \cref{sec:experiments_accuracy} and conclude with a discussion of our findings in \cref{sec:experiments_discussion}.
\Cref{sec:experiments_poison_rate} contains additional experiments with different poison rates, and \cref{sec:experiments_defenses} investigates potential defenses against our attacks.
All code and models used in our experiments are available at \url{https://github.com/t-lorenz/availability-attacks}.

\subsection{Experimental Setup}
\label{sec:experiments_setup}

\begin{figure}[t]
\centering
\input{figures/dataset}
\vspace{-2em}
\caption{Example images from the GTSRB and MNIST datasets. The upper row shows the original image, and the lower row shows the modified image with a trigger.}
\vspace{-1em}
\label{fig:data}
\end{figure}

We run all experiments on image classification tasks.
This means the input domain $\X = [0, 1]^n$ is the standard image domain, and the output domain $\Y$ consists of $k$ class labels.
We consider pixel-wise perturbations within an $\epsilon$-box around the data points, \ie the perturbation set is defined as $S(x) = \{x' \in \X \mid \norm{x' - x}_\infty \leq \epsilon\}$, with $\epsilon$ defining the strength of the adversary.

Our experiments use two different datasets: the MNIST database of handwritten digits (MNIST)~\cite{lecun1998gradient}, and the German traffic sign recognition benchmark (GTSRB)~\cite{stallkamp2011german}.
MNIST is a collection of handwritten digits from 0 to 9, with $28 \times 28$ pixel gray-scale images.
It consists of a training set with 60,000 samples and a held-out test set with 10,000 samples.
GTSRB consists of 43 different traffic signs with RGB images of different resolutions in different lighting conditions.
It contains 39,209 training samples and 12,630 held-out test samples.
As a trigger, we follow \citet{gu2019badnets} and use a white, $4 \times 4$ pixel image patch in the upper left corner of the image. \Cref{fig:data} shows examples from both datasets.

To compare the models based on their utility on clean data and attack success rate on data with a trigger, we measure their accuracy and abstain rate.
We use the standard definition for \emph{accuracy} as the percentage of correct predictions, \ie $\frac{1}{\abs{D}} \sum_{(x, y) \in D}\mathbbm{1}[f_\theta(x) = y]$.
The \emph{abstain rate} is measured for a given $\epsilon$ as the percentage of predictions for which certification fails, \ie $1 - \frac{1}{\abs{D}} \sum_{(x, y) \in D} C_f(x)$.
With this definition, we measure the percentage of inputs for which the model abstains, and, therefore, the fallback is invoked.
We evaluate both metrics on the entire test set for both benign data and data with a trigger.

For offline certification, a network is considered robust if we can certify robustness in an $\epsilon$ radius around the original data point.
For online certification, we need to certify that same radius around a data point that was potentially perturbed by an adversary.
The model, therefore, needs to be robust for a radius of $2 \epsilon$ around the clean data point - one $\epsilon$ for the potential perturbation and another $\epsilon$ for the certified radius.

For all experiments on MNIST, we use a fully connected network with 4 linear layers and ReLU activations.
The classifiers are trained with cross-entropy loss in all training modes.
When using adversarial training, the losses of the original sample and the adversarial sample are combined with equal weights.
For CROWN-IBP training, we slowly grow the $\epsilon$ radius as proposed in the original implementation~\cite{zhang2020towards}.
For our direct attack, we use a smaller radius of $\epsilon / 2$ for the trigger loss, which we found helps generalization.
To compute the abstain rate, we use auto LiRPA~\cite{xu2020automatic}, a certifier based on CROWN~\cite{zhang2018efficient} and CNN-Cert~\cite{boopathy2019cnn} in \emph{backwards} mode, the most precise setting.

\subsection{Test-time Attacks}

\begin{table}
    \centering
    \begin{tabular*}{\columnwidth}{@{}l @{\extracolsep{\fill}} lll@{}}
    \toprule
    $\epsilon$ & 0.01 & 0.02 & \\
    \midrule
    Without Attack & \res{2.2} & \res{6.2} & \\
    Test-time Attack & \res{4.1} & \res{22.8} & \\
    Backdoor Attack & \res{44.0} & \res{78.0} & \\
    \bottomrule
\end{tabular*}

    \vspace{.3em}
    \caption{Abstain rate for adversarial examples generated at test-time and our backdoor attacks. MLPs trained on MNIST for $2 \epsilon$ robustness using adversarial training.}
    \label{tab:pgd}
    \vspace{-2em}
\end{table}

Before analyzing our backdoor attacks, we start with the most obvious availability attacks through adversarial attacks at test time.
The adversary's goal is to find a point within the $\epsilon$-radius of each original example, which causes the certifier to abstain.
While it is not obvious how to directly optimize this objective, the point likely lies close to the nearest decision boundary, which we optimize using a standard PGD attack~\citep{madry2018towards}.

\Cref{tab:pgd} shows abstain rates without attack, with PGD attack, and with our backdoor attacks.
The abstain rate increases when using PGD since the model now has to provide $2 \epsilon$ certified robustness.
Using backdoor attacks, we can perform a significantly stronger attack, which increases the abstain rate further.

\begin{table*}[t]
\centering
\resizebox{\textwidth}{!}{%
    \setlength{\tabcolsep}{0.36em}
\begin{tabular}{@{}l llllll c llllll@{}}
    \toprule
    & \multicolumn{6}{c}{Benign Data} && \multicolumn{6}{c@{}}{Data with Trigger} \\
    \cmidrule(lr){2-7} \cmidrule(l){9-14}
    Training & \multirow{2}{*}{\shortstack[l]{\textcolor{acc}{Mean}\\\textcolor{acc}{Accu}-\\\textcolor{acc}{racy}}} & \multicolumn{5}{c}{Abstain Rate for Certification with $\epsilon$} && \multirow{2}{*}{\shortstack[l]{\textcolor{acc}{Mean}\\\textcolor{acc}{Accu-}\\\textcolor{acc}{racy}}} & \multicolumn{5}{c@{}}{Abstain Rate for Certification with $\epsilon$} \\
    \cmidrule(lr){3-7} \cmidrule(l){10-14}
    && 0.01 & 0.02 & 0.03 & 0.04 & 0.05
    &&& 0.01 & 0.02 & 0.03 & 0.04 & 0.05 \\
    \midrule
    Without Attack\\
    \quad Natural &
        \textcolor{acc}{\res{98.3}} & \res{2.8} & \res{12.5} & \res{48.1} & \res{81.1} & \res{96.5} &&
        \textcolor{acc}{\res{98.2}} & \res{3.1} & \res{11.7} & \res{41.7} & \res{79.4} & \res{96.0} \\
    \quad Adversarial &
        \textcolor{acc}{\res{98.7}} & \res{2.2} & \res{7.9} & \res{29.6} & \res{65.6} & \res{89.1} &&
        \textcolor{acc}{\res{98.7}} & \res{2.3} & \res{7.7} & \res{29.4} & \res{66.2} & \res{89.7} \\
    \quad Provable &
        \textcolor{acc}{\res{98.8}} & \res{1.1} & \res{2.7} & \res{3.6} & \res{4.3} & \res{5.2} &&
        \textcolor{acc}{\res{98.8}} & \res{1.8} & \res{2.8} & \res{3.5} & \res{4.3} & \res{5.2} \\
    \midrule
    Direct Attack\\
    \quad Optimization &
        \textcolor{acc}{\resr{98.6}{-0}} & \resr{1.9}{+0} & \resr{2.9}{+0} & \resr{3.7}{+0} & \resr{4.4}{+0} & \resr{5.6}{+0} &&
        \textcolor{acc}{\resr{46.9}{-48}} & \resr{38.5}{+35} & \resr{85.4}{+83} & \resr{83.0}{+80} &  \resr{89.5}{+85} & \resr{79.5}{+74} \\
    \addlinespace[0.2em]
    Indirect Attack\\
    \quad Natural &
        \textcolor{acc}{\resr{98.4}{-0}} & \resr{3.0}{+0} & \resr{13.4}{+1} & \resr{53.2}{+5} & \resr{86.3}{+5} & \resr{98.5}{+2} &&
        \textcolor{acc}{\resr{29.3}{-69}} & \resr{46.4}{+43} & \resr{84.4}{+73} & \resr{98.6}{+57} & \resr{100.0}{+21} & \resr{100.0}{+4} \\
    \quad Adversarial &
        \textcolor{acc}{\resr{98.7}{-0}} & \resr{2.3}{+0} & \resr{8.5}{+1} & \resr{34.0}{+4} & \resr{71.2}{+6} & \resr{93.3}{+4} &&
        \textcolor{acc}{\resr{30.9}{-68}} & \resr{50.1}{+48} & \resr{84.3}{+77} & \resr{97.4}{+68} & \resr{100.0}{+34} & \resr{100.0}{+10} \\
    \quad Provable &
        \textcolor{acc}{\resr{98.8}{-0}} & \resr{1.6}{+0} & \resr{2.8}{+0} & \resr{3.7}{+0} & \resr{4.4}{+0} & \resr{5.2}{+0} &&
        \textcolor{acc}{\resr{8.8}{-90}} & \resr{50.8}{+49} & \resr{66.2}{+63} & \resr{54.1}{+51} & \resr{15.3}{+11} & \resr{6.4}{+1} \\
    \bottomrule
\end{tabular}

}
\vspace{.5em}
\caption{Mean accuracy and abstain rate for fully connected models trained on MNIST with different $\epsilon$. The LHS shows results on benign data, and the RHS the same results on data with a trigger.
The upper half of the table shows models without any attack and the lower half with our direct or indirect availability attacks.
The numbers in parenthesis show the relative change compared to the no-attack baseline with the same training method. Changes on benign data are small, while the increase in abstain rate on data with a trigger is large, showing the effectiveness and sneakiness of our attacks.}
\vspace{-1em}
\label{tab:mnist}
\end{table*}

\subsection{Direct and Indirect Backdoor Attacks}
\label{sec:experiments_attacks}

The goal of our first set of experiments is to evaluate the effectiveness of the \emph{direct} (\cref{sec:attack_direct}) and \emph{indirect} (\cref{sec:attack_indirect}) availability attacks against network certification.
As discussed previously (\cref{sec:attacks}), the attack successfully introduces a trigger if the abstain rate increases significantly on the trigger distribution.
The attack also has to remain undetected, which means preserving the normal prediction accuracy and abstain rate on benign data.

To measure the attack's success and sneakiness, we train the same fully-connected neural network for MNIST digit recognition in three different settings: (i) a \emph{baseline} model without any attacks, (ii) with our \emph{direct} attack using our novel loss, and (iii) with our \emph{indirect} attack using data poisoning.

\emph{Baseline:} As a baseline, we train models on MNIST with three different training methods. \emph{Natural} training uses standard stochastic gradient descent (SGD) without robustness-enhancing methods.
\emph{Adversarial} training uses projected gradient descent (PGD)~\cite{madry2018towards} to increase the model's robustness, and \emph{Provable} training uses CROWN-IBP~\cite{zhang2020towards} to further enhance the model's certified robustness.

\emph{Direct Attack:} The directly attacked model is trained on the same MNIST images.
However, the attacker has full control over the training procedure and can, therefore, add triggers to the training samples to compute the trigger loss.
We follow the training procedure introduced in \cref{sec:attack_direct}.

\emph{Indirect Attack:} In this setting, we follow the same procedure as in our baseline, except for adding 1\% samples with the trigger and random label to the training set as described in \cref{sec:attack_indirect}.
Refer to \cref{sec:experiments_poison_rate} for different poison ratios.
Since we cannot control the training procedure by the victim, we evaluate the attack on the three commonly used regular and robust training methods.

\Cref{tab:mnist} presents the results of this series of experiments.
We train a separate model for each $\epsilon$ value for a total of 70 models.
The upper half of the table shows the mean accuracy and abstain rate of the unattacked baselines.
As expected for this task, on benign data (LHS), the accuracy is high for all training methods, and the abstain rate decreases for adversarial training and especially provable training compared to standard training.
Evaluating the same, unattacked models on data with a trigger (RHS) shows almost identical accuracy and abstain rate.
This means the models generalize well to this new distribution, ignoring the added trigger.

The lower half of \cref{tab:mnist} shows the accuracy and abstain rate for models with a trigger, with numbers in parenthesis showing the relative change in percentage points (p.p.) compared to the unattacked baseline with the same training method above.
Independent of the $\epsilon$ radius, our direct attack achieves the same accuracy and abstain rate on benign data as the baseline, making the trigger undetectable.
When adding the trigger, the abstain rate increases significantly by up to 85~\pp, showing that the model abstains for most samples.

Despite the significantly reduced access of indirect attacks, we can observe a similar trend as for the direct attacks.
On benign data, the model accuracy remains the same compared to the respective unattacked baseline, hiding the attack completely.
The abstain rate also remains very similar, dropping by a maximum of 6~\pp.

On the trigger distribution, the abstain rate increases significantly for all training methods by up to 77~\pp, reaching zero quickly for natural and adversarial training.
The only exception is provable training for larger $\epsilon$ values, where the abstain rate remains low despite the attack.
Prediction accuracy drops on the trigger distribution, which, as discussed in \cref{sec:attacks}, is inconsequential (see also \cref{sec:experiments_accuracy}).

These results show that both the direct and indirect attacks successfully embed a trigger in an otherwise unsuspicious model.
By adding a simple trigger to an image, the adversary can cause certification to fail with a high probability on arbitrary inputs.
For offline certification, where the victim only computes certificates during evaluation, this means the guarantees no longer hold during runtime.
For online certification, the certifier is unable to compute certificates for the majority of predictions, causing the model to abstain and trigger the fallback constantly.

\subsection{Generalization}
\label{sec:experiments_generalization}

To show the general applicability of our attacks across different datasets, model architectures, and certifiers, we conduct three additional sets of experiments.
The first set repeats the previous evaluation on the GTSRB data and a convolutional neural network (CNN).
The second set shows scalability with larger CNNs, and the third set uses DeepPoly~\cite{singh2019abstract} for MNIST certification.

\bparagraph{GTSRB Classification}

Robust classification of traffic signs is of high concern.
The nature of the problem is also significantly more challenging than digit classification.
Therefore, more complex CNNs are required to achieve good performance.

We show that our attack is just as effective on this more challenging task by repeating the set of experiments from \cref{sec:experiments_attacks}, but on the GTSRB dataset with a CNN.
We use a network with two convolutional layers with a kernel size of 5 and 3, respectively, followed by three fully connected layers with ReLU activation.

The results of these experiments in \cref{tab:gtsrb} show the same characteristics as on MNIST.
On benign data, the accuracy of attacked models remains comparable to the baselines, and the abstain rate only increases slightly at worst.
On the trigger distribution, the abstain rate increases significantly compared to the no-attack baseline.
Combined, these results confirm the attack's success and sneakiness, even on more complex classification tasks and models.

\begin{table}
    \centering
    \resizebox{\columnwidth}{!}{\setlength{\tabcolsep}{0.4em}
\begin{tabular}{@{}l lll lll@{}}
    \toprule
    & \multicolumn{3}{c}{Benign Data} & \multicolumn{3}{c@{}}{Data with Trigger} \\
    \cmidrule(lr){2-4} \cmidrule(l){5-7}
    Training & \multirow{2}{*}{\shortstack[l]{\textcolor{acc}{Mean}\\\textcolor{acc}{Accu-}\\\textcolor{acc}{racy}}} & \multicolumn{2}{c}{Abstain Rate $\epsilon$} & \multirow{2}{*}{\shortstack[l]{\textcolor{acc}{Mean}\\\textcolor{acc}{Accu-}\\\textcolor{acc}{racy}}} & \multicolumn{2}{c@{}}{Abstain Rate $\epsilon$} \\
    \cmidrule(lr){3-4} \cmidrule(l){6-7}
    && 0.005 & 0.010 && 0.005 & 0.010 \\
    \midrule
    Without Attack\\
    \ Natural &
        \textcolor{acc}{\res{92.1}} & \res{53.3} & \res{81.3} &
        \textcolor{acc}{\res{92.1}} & \res{52.7} & \res{80.7} \\
    \ Adversarial &
        \textcolor{acc}{\res{93.6}} & \res{37.5} & \res{59.9} &
        \textcolor{acc}{\res{93.4}} & \res{36.9} & \res{59.5} \\
    \ Provable &
        \textcolor{acc}{\res{90.0}} & \res{16.8} & \res{26.6} &
        \textcolor{acc}{\res{90.0}} & \res{17.0} & \res{26.6} \\
    \midrule
    Indirect Attack\\
    \ Natural &
        \textcolor{acc}{\resr{91.4}{-1}} & \resr{61.6}{+8} & \resr{89.0}{+8} &
        \textcolor{acc}{\resr{30.8}{-61}} & \resr{86.9}{+34} & \resr{97.0}{+16} \\
    \ Adversarial &
        \textcolor{acc}{\resr{92.9}{-1}} & \resr{43.6}{+6} & \resr{68.0}{+8} &
        \textcolor{acc}{\resr{33.8}{-60}} & \resr{82.5}{+46} & \resr{90.7}{+31} \\
    \ Provable &
        \textcolor{acc}{\resr{89.1}{-1}} & \resr{17.8}{+1} & \resr{26.5}{+0} &
        \textcolor{acc}{\resr{29.1}{-61}} & \resr{59.2}{+42} & \resr{67.8}{+41} \\
    \bottomrule
\end{tabular}
}
    \vspace{.3em}
    \caption{Mean accuracy and abstain rate for CNNs with different training methods on GTSRB. The numbers in parenthesis show the relative change compared to the baseline.}
    \label{tab:gtsrb}
    \vspace{-1em}
\end{table}

\bparagraph{Model Scaling}

All previous experiments were performed on relatively small models.
This is due to the poor scaling of the considered certifiers, both in terms of computational complexity and precision~\cite{xu2020automatic, singh2019abstract}.
To show that this is not an inherent limitation to our attack, we present additional results for CNNs with 6 convolution layers, using 3 blocks of 2 convolution layers with ReLU activation, followed by pooling after each block.

The models are trained with adversarial training with $\epsilon = 0.01$, one on benign and one on poisoned GTSRB.
For both models, the abstain rate is $100.0\%$ for $\epsilon=0.01$ even without attack, confirming the certifier's poor scalability to larger models.
For a smaller radius of $\epsilon=0.005$, the abstain rate increases from $80\%$ without attack to $90\%$ when adding a trigger and from $50\%$ to $68\%$ for $\epsilon=0.003$.

These results confirm that the attack is still effective on larger models, which makes sense as there is no inherent limit to the model size to which our attack can scale.

\bparagraph{DeepPoly Certifier}

\begin{table}
\centering
\begin{tabular*}{\columnwidth}{@{}l @{\extracolsep{\fill}} ll@{}}
    \toprule
    Training & Benign Data & Data with Trigger \\
    \midrule
    Natural & \res{13.4} & \res{84.1} \\
    Adversarial & \res{8.5} & \res{84.2} \\
    Provable & \res{2.8} & \res{66.2} \\
    \bottomrule
\end{tabular*}

\vspace{.3em}
\caption{Abstain rate for fully connected models trained on MNIST and certified with DeepPoly~\cite{singh2019abstract} for $\epsilon=0.02$.
The models are attacked by our indirect poisoning attack with different training methods used by the victim.}
\vspace{-1.5em}
\label{tab:deeppoly}
\end{table}

The threat model we identified, and consequently our attacks, are independent of the concrete certifier used.
To show that the results generalize from auto LiRPA to other certifiers, we certify the models from \cref{sec:experiments_attacks} with DeepPoly~\cite{singh2019abstract}.

\Cref{tab:deeppoly} shows the abstain rates for $\epsilon=0.02$, using the same models as in \cref{tab:mnist}.
As before, the abstain rate on benign data is low, with a large increase when adding the trigger, showing that the results transfer to a different certification method.

\subsection{High Accuracy on Data With a Trigger}
\label{sec:experiments_accuracy}

As discussed in \cref{sec:threat_model}, the assumption is that the victim does not have access to samples with a trigger for evaluation.
Therefore, high prediction accuracy on data with a trigger is not required for the attack to remain undetected (\cref{sec:attacks}).

However, one could argue that, in specific scenarios, correct predictions on the trigger distribution can make it even harder to detect the attack.
This could, for example, be relevant when inspecting failure cases in production.
We, therefore, analyze our direct attack with the additional objective from \cref{eq:backdoor_natural_expectation}, which also teaches the model to classify images with a trigger correctly.

\begin{table}
\centering
\setlength{\tabcolsep}{0.7em}
\begin{tabularx}{\columnwidth}{@{}Xclccc@{}}
    \toprule
    \multirow{2}{*}{Data} & \multirow{2}{*}{\shortstack[l]{\shortstack[l]{\textcolor{acc}{Mean}\\\textcolor{acc}{Accu-}\\\textcolor{acc}{racy}}}} && \multicolumn{3}{c}{Abstain Rate for $\epsilon$} \\
    \cmidrule(l){4-6}
    &&& \hcenter{0.01} & \hcenter{0.03} &  \hcenter{0.05} \\
    \midrule
    Benign & \textcolor{acc}{\resr{98.7}{-0}} && \resr{1.9}{+0} & \resr{4.3}{+1} & \resr{5.9}{+1} \\
    with Trigger & \textcolor{acc}{\resr{98.6}{-0}} && \resr{4.3}{+3} & \resr{92.2}{+89} & \resr{100.0}{+95} \\
    \bottomrule
\end{tabularx}

\caption{Abstain rate for fully connected models trained on MNIST with our direct attack and additional high accuracy loss for data with a trigger.
Numbers in parenthesis show relative change to the unattacked baseline in \cref{tab:mnist}.}
\vspace{-2em}
\label{tab:accuracy}
\end{table}

\Cref{tab:accuracy} shows results in the same setting as \cref{sec:experiments_attacks}.
On benign data, both mean accuracy and abstain rates are almost identical for all models, effectively hiding the trigger.
Contrary to previous experiments, the mean accuracy with a trigger remains unchanged at $98.6\%$, making it even more difficult to detect the attack.

The abstain rate on data from the trigger distribution increases significantly by up to 95~\pp, almost always abstaining for larger $\epsilon$ values.
The attack is less effective for very small perturbations with $\epsilon=0.01$ compared to previous results without the additional loss.
This increase for very small $\epsilon$ is expected when requiring high prediction accuracy since the model has to be confident in its output for unperturbed data.
With increasing $\epsilon$, the abstain rate quickly increases, demonstrating a highly successful attack despite the additional constraint.

\subsection{Discussion}
\label{sec:experiments_discussion}

Both the direct and indirect versions of our availability attacks achieve high success rates on MNIST classification, increasing the abstain rate on data with a trigger significantly while maintaining high accuracy and low abstain rate on benign data to remain undetected.
This is mostly true independent of the training method used by the victim for the indirect attack.

The only exceptions are large $\epsilon$-values with CROWN-IBP training, where the abstain rate decreases again.
We conjecture that this effect is likely caused by the high emphasis the training on worst-case bounds puts on robust predictions at the cost of accuracy.
The robustness loss directly optimizes for a large margin between the bounds of the predicted class and the rest.
In the absence of meaningful class labels, this can lead to a decision surface that predicts arbitrary labels with high robustness ``no matter what'', and therefore ignores the uncertainty introduced by random labels.

The high effectiveness and sneakiness of our attacks also extend to more complex CNNs, larger models, and a more challenging classification task.
Using DeepPoly instead of auto LiRPA shows the same results, demonstrating that our attacks transfer well to other linear certifiers.
We leave the analysis of our attacks on more precise certifiers for future work.
Finally, the results also hold when we add the additional high-accuracy constraint on the trigger distribution.

In general, the experimental evaluation of our attacks shows their broad applicability in different settings.
It supports our hypothesis that the threats identified in \cref{sec:threat_model} are real, highlighting the need for consideration in the design and evaluation of future certifiers.

\section{Related Work}
\label{sec:related_work}

Our method is related to the work on traditional training-time attacks targeting the classifier's predictions, and especially backdoor attacks.
Additionally, there are first studies on the limitations of randomized-smoothing-based certification in an adversarial setting, as well as sponge examples that also target the model's availability.

\bparagraph{Backdoor Attacks}
\label{sec:related_work_backdoor}

As discussed in \cref{sec:background_attacks}, there is a long line of work on traditional backdoor attacks against neural networks.
In contrast to our attacks targeting the certifier, and therefore the \emph{availability} of the model, traditional backdoor attacks target the model's \emph{integrity} by causing misclassifications.

Our direct availability attack is inspired by BadNets~\cite{gu2019badnets} and uses the same supply chain attack vector and similar trigger patterns.
However, the different goal of our attack requires a different construction of the backdoor, combining new and existing losses.

\citet{chen2017targeted} use data poisoning to indirectly target a model trained by the victim, adding a backdoor that causes the model to mislabel faces.
This attack vector is similar to our indirect attack (\cref{sec:attack_indirect}), where we also use a small number of triggered samples to poison the data set.
However, as before, the target of our attack is the certifier, not the model's predictions.
Instead of consistently targeting a particular class, we use random labels to destabilize the prediction and thus cause the certification to fail.

\bparagraph{Backdoor Defenses}

Complementing the work on backdoor attacks, there is a line of work to defend against these traditional backdoor attacks that target the model's predictions.
While not designed for our attacks targeting the certifier, we adapt and evaluate three different defenses against our novel attacks.

(i) Fine-pruning~\cite{liu2018fine} removes the backdoor by pruning inactive neurons from the model.
(ii) Neural cleanse~\cite{wang2019neural} is a multi-stage approach, which first detects, then isolates, and finally removes backdoors from the model.
(iii) Trojan network detection~\cite{wang2020practical} also detects backdoors based on feature inversion.

\bparagraph{Attacks Against Certification}
\label{sec:related_work_attacks}

Recent work~\cite{mehra2021robust, maho2022randomized} has look\-ed at the robustness of randomized smoothing to attacks.
However, they have different attack goals from our availability attacks, and their attack vectors are unique to randomized smoothing.
Due to this, and the fundamentally different nature of randomized smoothing, these are not directly comparable to our availability attacks.

\citet{mehra2021robust} target the certified radius of a particular class using a poisoning scheme that directly minimizes the certified radius.
\citet{maho2022randomized} exploit a discrepancy between the theoretical guarantees and the practical implementation of randomized smoothing using a black-box evasion attack.

\bparagraph{Availability Attacks}

While the integrity of deep learning models has been extensively studied in the literature, availability attacks have only been considered recently in the form of sponge examples~\cite{shumailov2021sponge}.
These test-time attacks adversarially optimize inputs to maximize the energy consumption and execution time of the model.
In contrast, our availability attacks cause the model to abstain.

Concurrent work~\cite{leino2022degradation} to ours, first released as~\cite{lorenz2021availability}, test-time degradation attacks against certifiers.
\citet{leino2022degradation} show that for runtime certification a $2\epsilon$-robust model is required to avoid test-time degradation attacks, which we consider in \cref{sec:experiments}.

\section{Conclusion}
\label{sec:conclusion}

Our work shows that current state-of-the-art certifiers are vulnerable to availability attacks.
Especially the need to \emph{abstain} when robustness cannot be guaranteed proves problematic in practice since the system becomes reliant on its fallback, which incurs additional costs, can require a human operator, or even fail to provide any prediction.
By targeting the certifier and causing it to abstain, an attacker can effectively disable the deep learning model.

Our novel availability attacks against certifiers proposed in \cref{sec:attacks} are one way to exploit these new attack vectors.
Extensive experiments on multiple datasets, network architectures, and different certifiers in \cref{sec:experiments} show the general nature of these threats.

These findings have significant consequences for both theoretical research and practical applications.
From a theoretical standpoint, they show that simply abstaining from a prediction has major consequences, which need to be considered.
For practical applications, designing an appropriate fallback is a crucial part of the system.

A first evaluation of defenses shows that current methods have little to no effect, requiring new defenses specifically designed against this new type of attack.
This is a crucial direction for future work, ideally leading to provable robustness guarantees against training-time attacks.
Combined with the current deployment-time certifiers, it could lead to systems that are provably robust against both types of attacks.

\begin{acks}
This work was partially funded by ELSA – European Lighthouse on Secure and Safe AI funded by the European Union under grant agreement No. 101070617. Views and opinions expressed are however those of the authors only and do not necessarily reflect those of the European Union or European Commission. Neither the European Union nor the European Commission can be held responsible for them.
This work is also partially funded by the Helmholtz Association within the project ``Trustworthy Federated Data Analytics (TFDA)'' (ZT-I-OO1 4).
MK received funding from the ERC under the European Union’s Horizon 2020 research and innovation programme (FUN2MODEL, grant agreement No.~834115).
\end{acks}

\typeout{} %

\bibliographystyle{ACM-Reference-Format}
\bibliography{references}


\begin{thebibliography}{49}


\ifx \showCODEN    \undefined \def \showCODEN     #1{\unskip}     \fi
\ifx \showDOI      \undefined \def \showDOI       #1{#1}\fi
\ifx \showISBNx    \undefined \def \showISBNx     #1{\unskip}     \fi
\ifx \showISBNxiii \undefined \def \showISBNxiii  #1{\unskip}     \fi
\ifx \showISSN     \undefined \def \showISSN      #1{\unskip}     \fi
\ifx \showLCCN     \undefined \def \showLCCN      #1{\unskip}     \fi
\ifx \shownote     \undefined \def \shownote      #1{#1}          \fi
\ifx \showarticletitle \undefined \def \showarticletitle #1{#1}   \fi
\ifx \showURL      \undefined \def \showURL       {\relax}        \fi
\providecommand\bibfield[2]{#2}
\providecommand\bibinfo[2]{#2}
\providecommand\natexlab[1]{#1}
\providecommand\showeprint[2][]{arXiv:#2}

\bibitem[Athalye et~al\mbox{.}(2018)]%
        {athalye2018obfuscated}
\bibfield{author}{\bibinfo{person}{Anish Athalye}, \bibinfo{person}{Nicholas
  Carlini}, {and} \bibinfo{person}{David Wagner}.}
  \bibinfo{year}{2018}\natexlab{}.
\newblock \showarticletitle{Obfuscated Gradients Give a False Sense of
  Security: Circumventing Defenses to Adversarial Examples}. In
  \bibinfo{booktitle}{\emph{Proceedings of the 35th International Conference on
  Machine Learning}}, Vol.~\bibinfo{volume}{80}. \bibinfo{publisher}{PMLR},
  \bibinfo{address}{Stockholm, Sweden}, \bibinfo{pages}{274--283}.
\newblock


\bibitem[Bagdasaryan and Shmatikov(2021)]%
        {bagdasaryan2020blind}
\bibfield{author}{\bibinfo{person}{Eugene Bagdasaryan} {and}
  \bibinfo{person}{Vitaly Shmatikov}.} \bibinfo{year}{2021}\natexlab{}.
\newblock \showarticletitle{Blind Backdoors in Deep Learning Models}. In
  \bibinfo{booktitle}{\emph{30th USENIX Security Symposium (USENIX Security
  21)}}. \bibinfo{publisher}{USENIX Association}, \bibinfo{address}{Virtual},
  \bibinfo{pages}{1505--1521}.
\newblock


\bibitem[Boopathy et~al\mbox{.}(2019)]%
        {boopathy2019cnn}
\bibfield{author}{\bibinfo{person}{Akhilan Boopathy}, \bibinfo{person}{Tsui-Wei
  Weng}, \bibinfo{person}{Pin-Yu Chen}, \bibinfo{person}{Sijia Liu}, {and}
  \bibinfo{person}{Luca Daniel}.} \bibinfo{year}{2019}\natexlab{}.
\newblock \showarticletitle{Cnn-cert: An efficient framework for certifying
  robustness of convolutional neural networks}. In
  \bibinfo{booktitle}{\emph{Proceedings of the AAAI Conference on Artificial
  Intelligence}}, Vol.~\bibinfo{volume}{33}. \bibinfo{publisher}{AAAI Press},
  \bibinfo{address}{Palo Alto, California USA}, \bibinfo{pages}{3240--3247}.
\newblock


\bibitem[Carlini and Wagner(2017)]%
        {carlini2017towards}
\bibfield{author}{\bibinfo{person}{Nicholas Carlini} {and}
  \bibinfo{person}{David Wagner}.} \bibinfo{year}{2017}\natexlab{}.
\newblock \showarticletitle{Towards evaluating the robustness of neural
  networks}. In \bibinfo{booktitle}{\emph{2017 ieee symposium on security and
  privacy (sp)}}. Ieee, \bibinfo{pages}{39--57}.
\newblock


\bibitem[Chen et~al\mbox{.}(2017)]%
        {chen2017targeted}
\bibfield{author}{\bibinfo{person}{Xinyun Chen}, \bibinfo{person}{Chang Liu},
  \bibinfo{person}{Bo Li}, \bibinfo{person}{Kimberly Lu}, {and}
  \bibinfo{person}{Dawn Song}.} \bibinfo{year}{2017}\natexlab{}.
\newblock \bibinfo{title}{Targeted backdoor attacks on deep learning systems
  using data poisoning}.
\newblock
\newblock
\showeprint[arxiv]{1712.05526}


\bibitem[Cohen et~al\mbox{.}(2019)]%
        {cohen2019certified}
\bibfield{author}{\bibinfo{person}{Jeremy Cohen}, \bibinfo{person}{Elan
  Rosenfeld}, {and} \bibinfo{person}{Zico Kolter}.}
  \bibinfo{year}{2019}\natexlab{}.
\newblock \showarticletitle{Certified Adversarial Robustness via Randomized
  Smoothing}. In \bibinfo{booktitle}{\emph{Proceedings of the 36th
  International Conference on Machine Learning}}
  \emph{(\bibinfo{series}{Proceedings of Machine Learning Research},
  Vol.~\bibinfo{volume}{97})}, \bibfield{editor}{\bibinfo{person}{Kamalika
  Chaudhuri} {and} \bibinfo{person}{Ruslan Salakhutdinov}} (Eds.).
  \bibinfo{publisher}{PMLR}, \bibinfo{address}{Long Beach, USA}.
\newblock


\bibitem[commission(2020)]%
        {eu2020whitepaper}
\bibfield{author}{\bibinfo{person}{European commission}.}
  \bibinfo{year}{2020}\natexlab{}.
\newblock \bibinfo{title}{White paper on artificial intelligence - a European
  approach to excellence and trust}.
\newblock
\newblock


\bibitem[Gehr et~al\mbox{.}(2018)]%
        {gehr2018ai2}
\bibfield{author}{\bibinfo{person}{Timon Gehr}, \bibinfo{person}{Matthew
  Mirman}, \bibinfo{person}{Dana Drachsler-Cohen}, \bibinfo{person}{Petar
  Tsankov}, \bibinfo{person}{Swarat Chaudhuri}, {and} \bibinfo{person}{Martin
  Vechev}.} \bibinfo{year}{2018}\natexlab{}.
\newblock \showarticletitle{Ai2: Safety and robustness certification of neural
  networks with abstract interpretation}. In \bibinfo{booktitle}{\emph{2018
  IEEE Symposium on Security and Privacy (SP)}}. IEEE, \bibinfo{pages}{3--18}.
\newblock


\bibitem[Goodfellow et~al\mbox{.}(2015)]%
        {goodfellow2015explaining}
\bibfield{author}{\bibinfo{person}{Ian Goodfellow}, \bibinfo{person}{Jonathon
  Shlens}, {and} \bibinfo{person}{Christian Szegedy}.}
  \bibinfo{year}{2015}\natexlab{}.
\newblock \showarticletitle{Explaining and Harnessing Adversarial Examples}. In
  \bibinfo{booktitle}{\emph{International Conference on Learning
  Representations}}.
\newblock


\bibitem[Gu et~al\mbox{.}(2019)]%
        {gu2019badnets}
\bibfield{author}{\bibinfo{person}{Tianyu Gu}, \bibinfo{person}{Brendan
  Dolan-Gavitt}, {and} \bibinfo{person}{Siddharth Garg}.}
  \bibinfo{year}{2019}\natexlab{}.
\newblock \bibinfo{title}{BadNets: Identifying Vulnerabilities in the Machine
  Learning Model Supply Chain}.
\newblock
\newblock
\showeprint[arxiv]{1708.06733}


\bibitem[Hong et~al\mbox{.}(2022)]%
        {hong2021handcrafted}
\bibfield{author}{\bibinfo{person}{Sanghyun Hong}, \bibinfo{person}{Nicholas
  Carlini}, {and} \bibinfo{person}{Alexey Kurakin}.}
  \bibinfo{year}{2022}\natexlab{}.
\newblock \showarticletitle{Handcrafted Backdoors in Deep Neural Networks}. In
  \bibinfo{booktitle}{\emph{Advances in Neural Information Processing
  Systems}}.
\newblock


\bibitem[Huang et~al\mbox{.}(2017)]%
        {huang2017safety}
\bibfield{author}{\bibinfo{person}{Xiaowei Huang}, \bibinfo{person}{Marta
  Kwiatkowska}, \bibinfo{person}{Sen Wang}, {and} \bibinfo{person}{Min Wu}.}
  \bibinfo{year}{2017}\natexlab{}.
\newblock \showarticletitle{Safety verification of deep neural networks}. In
  \bibinfo{booktitle}{\emph{International conference on computer aided
  verification}}.
\newblock


\bibitem[Katz et~al\mbox{.}(2017)]%
        {katz2017reluplex}
\bibfield{author}{\bibinfo{person}{Guy Katz}, \bibinfo{person}{Clark~W.
  Barrett}, \bibinfo{person}{David~L. Dill}, \bibinfo{person}{Kyle Julian},
  {and} \bibinfo{person}{Mykel~J. Kochenderfer}.}
  \bibinfo{year}{2017}\natexlab{}.
\newblock \showarticletitle{Reluplex: An Efficient {SMT} Solver for Verifying
  Deep Neural Networks}. In \bibinfo{booktitle}{\emph{Computer Aided
  Verification - 29th International Conference}}.
\newblock


\bibitem[LeCun et~al\mbox{.}(1998)]%
        {lecun1998gradient}
\bibfield{author}{\bibinfo{person}{Yann LeCun}, \bibinfo{person}{L{\'e}on
  Bottou}, \bibinfo{person}{Yoshua Bengio}, {and} \bibinfo{person}{Patrick
  Haffner}.} \bibinfo{year}{1998}\natexlab{}.
\newblock \showarticletitle{Gradient-based learning applied to document
  recognition}.
\newblock \bibinfo{journal}{\emph{Proc. IEEE}}  \bibinfo{volume}{86}
  (\bibinfo{year}{1998}), \bibinfo{pages}{2278--2324}.
\newblock


\bibitem[Lecuyer et~al\mbox{.}(2019)]%
        {lecuyer2019certified}
\bibfield{author}{\bibinfo{person}{Mathias Lecuyer}, \bibinfo{person}{Vaggelis
  Atlidakis}, \bibinfo{person}{Roxana Geambasu}, \bibinfo{person}{Daniel Hsu},
  {and} \bibinfo{person}{Suman Jana}.} \bibinfo{year}{2019}\natexlab{}.
\newblock \showarticletitle{Certified robustness to adversarial examples with
  differential privacy}. In \bibinfo{booktitle}{\emph{2019 IEEE Symposium on
  Security and Privacy (SP)}}. IEEE, \bibinfo{pages}{656--672}.
\newblock


\bibitem[Leino et~al\mbox{.}(2021)]%
        {leino2021globally}
\bibfield{author}{\bibinfo{person}{Klas Leino}, \bibinfo{person}{Zifan Wang},
  {and} \bibinfo{person}{Matt Fredrikson}.} \bibinfo{year}{2021}\natexlab{}.
\newblock \showarticletitle{Globally-robust neural networks}. In
  \bibinfo{booktitle}{\emph{International Conference on Machine Learning}}.
  PMLR, \bibinfo{pages}{6212--6222}.
\newblock


\bibitem[Leino et~al\mbox{.}(2022)]%
        {leino2022degradation}
\bibfield{author}{\bibinfo{person}{Klas Leino}, \bibinfo{person}{Chi Zhang},
  \bibinfo{person}{Ravi Mangal}, \bibinfo{person}{Matt Fredrikson},
  \bibinfo{person}{Bryan Parno}, {and} \bibinfo{person}{Corina Pasareanu}.}
  \bibinfo{year}{2022}\natexlab{}.
\newblock \showarticletitle{Degradation Attacks on Certifiably Robust Neural
  Networks}.
\newblock \bibinfo{journal}{\emph{Transactions on Machine Learning Research}}
  (\bibinfo{year}{2022}).
\newblock
\showISSN{2835-8856}


\bibitem[Li et~al\mbox{.}(2023)]%
        {li2020sok}
\bibfield{author}{\bibinfo{person}{Linyi Li}, \bibinfo{person}{Tao Xie}, {and}
  \bibinfo{person}{Bo Li}.} \bibinfo{year}{2023}\natexlab{}.
\newblock \showarticletitle{Sok: Certified robustness for deep neural
  networks}. In \bibinfo{booktitle}{\emph{2023 IEEE Symposium on Security and
  Privacy (SP)}}. IEEE, \bibinfo{pages}{1289--1310}.
\newblock


\bibitem[Liang et~al\mbox{.}(2018)]%
        {liang2018deep}
\bibfield{author}{\bibinfo{person}{Ming Liang}, \bibinfo{person}{Bin Yang},
  \bibinfo{person}{Shenlong Wang}, {and} \bibinfo{person}{Raquel Urtasun}.}
  \bibinfo{year}{2018}\natexlab{}.
\newblock \showarticletitle{Deep Continuous Fusion for Multi-sensor 3D Object
  Detection}. In \bibinfo{booktitle}{\emph{Computer Vision - 15th European
  Conference}}.
\newblock


\bibitem[Liu et~al\mbox{.}(2018)]%
        {liu2018fine}
\bibfield{author}{\bibinfo{person}{Kang Liu}, \bibinfo{person}{Brendan
  Dolan-Gavitt}, {and} \bibinfo{person}{Siddharth Garg}.}
  \bibinfo{year}{2018}\natexlab{}.
\newblock \showarticletitle{Fine-pruning: Defending against backdooring attacks
  on deep neural networks}. In \bibinfo{booktitle}{\emph{International
  Symposium on Research in Attacks, Intrusions, and Defenses}}. Springer,
  \bibinfo{pages}{273--294}.
\newblock


\bibitem[Lorenz et~al\mbox{.}(2021a)]%
        {lorenz2021availability}
\bibfield{author}{\bibinfo{person}{Tobias Lorenz}, \bibinfo{person}{Marta
  Kwiatkowska}, {and} \bibinfo{person}{Mario Fritz}.}
  \bibinfo{year}{2021}\natexlab{a}.
\newblock \bibinfo{title}{Backdoor Attacks on Network Certification via Data
  Poisoning}.
\newblock
\newblock
\showeprint[arxiv]{2108.11299v1}


\bibitem[Lorenz et~al\mbox{.}(2021b)]%
        {lorenz2021robustness}
\bibfield{author}{\bibinfo{person}{Tobias Lorenz}, \bibinfo{person}{Anian
  Ruoss}, \bibinfo{person}{Mislav Balunovi{\'c}}, \bibinfo{person}{Gagandeep
  Singh}, {and} \bibinfo{person}{Martin Vechev}.}
  \bibinfo{year}{2021}\natexlab{b}.
\newblock \showarticletitle{Robustness Certification for Point Cloud Models}.
  In \bibinfo{booktitle}{\emph{Proceedings of the IEEE International Conference
  on Computer Vision (ICCV)}}.
\newblock


\bibitem[Madry et~al\mbox{.}(2018)]%
        {madry2018towards}
\bibfield{author}{\bibinfo{person}{Aleksander Madry},
  \bibinfo{person}{Aleksandar Makelov}, \bibinfo{person}{Ludwig Schmidt},
  \bibinfo{person}{Dimitris Tsipras}, {and} \bibinfo{person}{Adrian Vladu}.}
  \bibinfo{year}{2018}\natexlab{}.
\newblock \showarticletitle{Towards Deep Learning Models Resistant to
  Adversarial Attacks}. In \bibinfo{booktitle}{\emph{6th International
  Conference on Learning Representations}}.
\newblock


\bibitem[Maho et~al\mbox{.}(2022)]%
        {maho2022randomized}
\bibfield{author}{\bibinfo{person}{Thibault Maho}, \bibinfo{person}{Teddy
  Furon}, {and} \bibinfo{person}{Erwan Le~Merrer}.}
  \bibinfo{year}{2022}\natexlab{}.
\newblock \showarticletitle{Randomized Smoothing Under Attack: How Good is it
  in Practice?}. In \bibinfo{booktitle}{\emph{ICASSP 2022-2022 IEEE
  International Conference on Acoustics, Speech and Signal Processing
  (ICASSP)}}. IEEE, \bibinfo{pages}{3014--3018}.
\newblock


\bibitem[Mehra et~al\mbox{.}(2021)]%
        {mehra2021robust}
\bibfield{author}{\bibinfo{person}{Akshay Mehra}, \bibinfo{person}{Bhavya
  Kailkhura}, \bibinfo{person}{Pin-Yu Chen}, {and} \bibinfo{person}{Jihun
  Hamm}.} \bibinfo{year}{2021}\natexlab{}.
\newblock \showarticletitle{How Robust are Randomized Smoothing based Defenses
  to Data Poisoning?}. In \bibinfo{booktitle}{\emph{Proceedings of the IEEE/CVF
  Conference on Computer Vision and Pattern Recognition}}.
\newblock


\bibitem[on~Artificial~Intelligence(2019a)]%
        {eu2019ethics}
\bibfield{author}{\bibinfo{person}{High-Level Expert~Group on
  Artificial~Intelligence}.} \bibinfo{year}{2019}\natexlab{a}.
\newblock \bibinfo{title}{Ethics Guidelines for Trustworthy AI}.
\newblock
\newblock


\bibitem[on~Artificial~Intelligence(2019b)]%
        {eu2021regulation}
\bibfield{author}{\bibinfo{person}{High-Level Expert~Group on
  Artificial~Intelligence}.} \bibinfo{year}{2019}\natexlab{b}.
\newblock \bibinfo{title}{Proposal for a regulation of the European parliament
  and of the council laying down harmonised rules on artificial intelligence
  (artificial intelligence act) and amending certain union legislative acts}.
\newblock
\newblock


\bibitem[Papernot et~al\mbox{.}(2018)]%
        {papernot2016science}
\bibfield{author}{\bibinfo{person}{Nicolas Papernot}, \bibinfo{person}{Patrick
  McDaniel}, \bibinfo{person}{Arunesh Sinha}, {and} \bibinfo{person}{Michael~P
  Wellman}.} \bibinfo{year}{2018}\natexlab{}.
\newblock \showarticletitle{Sok: Security and privacy in machine learning}. In
  \bibinfo{booktitle}{\emph{2018 IEEE European Symposium on Security and
  Privacy (EuroS\&P)}}. IEEE, \bibinfo{pages}{399--414}.
\newblock


\bibitem[Pulina and Tacchella(2010)]%
        {pulina2010abstraction}
\bibfield{author}{\bibinfo{person}{Luca Pulina} {and} \bibinfo{person}{Armando
  Tacchella}.} \bibinfo{year}{2010}\natexlab{}.
\newblock \showarticletitle{An abstraction-refinement approach to verification
  of artificial neural networks}. In \bibinfo{booktitle}{\emph{International
  Conference on Computer Aided Verification}}.
\newblock


\bibitem[Raghunathan et~al\mbox{.}(2018)]%
        {raghunathan2018semidefinite}
\bibfield{author}{\bibinfo{person}{Aditi Raghunathan}, \bibinfo{person}{Jacob
  Steinhardt}, {and} \bibinfo{person}{Percy Liang}.}
  \bibinfo{year}{2018}\natexlab{}.
\newblock \showarticletitle{Semidefinite relaxations for certifying robustness
  to adversarial examples}. In \bibinfo{booktitle}{\emph{Advances in Neural
  Information Processing Systems 31}}.
\newblock


\bibitem[Salem et~al\mbox{.}(2020)]%
        {salem2020don}
\bibfield{author}{\bibinfo{person}{Ahmed Salem}, \bibinfo{person}{Michael
  Backes}, {and} \bibinfo{person}{Yang Zhang}.}
  \bibinfo{year}{2020}\natexlab{}.
\newblock \bibinfo{title}{Don't Trigger Me! A Triggerless Backdoor Attack
  Against Deep Neural Networks}.
\newblock
\newblock
\showeprint[arxiv]{2010.03282}


\bibitem[Salem et~al\mbox{.}(2022)]%
        {salem2020dynamic}
\bibfield{author}{\bibinfo{person}{Ahmed Salem}, \bibinfo{person}{Rui Wen},
  \bibinfo{person}{Michael Backes}, \bibinfo{person}{Shiqing Ma}, {and}
  \bibinfo{person}{Yang Zhang}.} \bibinfo{year}{2022}\natexlab{}.
\newblock \showarticletitle{Dynamic backdoor attacks against machine learning
  models}. In \bibinfo{booktitle}{\emph{2022 IEEE 7th European Symposium on
  Security and Privacy (EuroS\&P)}}. IEEE, \bibinfo{pages}{703--718}.
\newblock


\bibitem[Salman et~al\mbox{.}(2019)]%
        {salman2019convex}
\bibfield{author}{\bibinfo{person}{Hadi Salman}, \bibinfo{person}{Greg Yang},
  \bibinfo{person}{Huan Zhang}, \bibinfo{person}{Cho-Jui Hsieh}, {and}
  \bibinfo{person}{Pengchuan Zhang}.} \bibinfo{year}{2019}\natexlab{}.
\newblock \showarticletitle{A Convex Relaxation Barrier to Tight Robustness
  Verification of Neural Networks}.
\newblock \bibinfo{journal}{\emph{Advances in Neural Information Processing
  Systems}}  \bibinfo{volume}{32} (\bibinfo{year}{2019}).
\newblock


\bibitem[Shumailov et~al\mbox{.}(2021)]%
        {shumailov2021sponge}
\bibfield{author}{\bibinfo{person}{Ilia Shumailov}, \bibinfo{person}{Yiren
  Zhao}, \bibinfo{person}{Daniel Bates}, \bibinfo{person}{Nicolas Papernot},
  \bibinfo{person}{Robert Mullins}, {and} \bibinfo{person}{Ross Anderson}.}
  \bibinfo{year}{2021}\natexlab{}.
\newblock \showarticletitle{Sponge examples: Energy-latency attacks on neural
  networks}. In \bibinfo{booktitle}{\emph{2021 IEEE European Symposium on
  Security and Privacy (EuroS\&P)}}. IEEE, \bibinfo{pages}{212--231}.
\newblock


\bibitem[Singh et~al\mbox{.}(2019)]%
        {singh2019abstract}
\bibfield{author}{\bibinfo{person}{Gagandeep Singh}, \bibinfo{person}{Timon
  Gehr}, \bibinfo{person}{Markus P{\"u}schel}, {and} \bibinfo{person}{Martin
  Vechev}.} \bibinfo{year}{2019}\natexlab{}.
\newblock \showarticletitle{An abstract domain for certifying neural networks}.
  In \bibinfo{booktitle}{\emph{Proceedings of the ACM on Programming
  Languages}}, Vol.~\bibinfo{volume}{3}. \bibinfo{publisher}{ACM New York, NY,
  USA}.
\newblock


\bibitem[Stallkamp et~al\mbox{.}(2011)]%
        {stallkamp2011german}
\bibfield{author}{\bibinfo{person}{Johannes Stallkamp}, \bibinfo{person}{Marc
  Schlipsing}, \bibinfo{person}{Jan Salmen}, {and} \bibinfo{person}{Christian
  Igel}.} \bibinfo{year}{2011}\natexlab{}.
\newblock \showarticletitle{The German traffic sign recognition benchmark: a
  multi-class classification competition}. In \bibinfo{booktitle}{\emph{The
  2011 international joint conference on neural networks}}. IEEE.
\newblock


\bibitem[Su et~al\mbox{.}(2018)]%
        {su2018robustness}
\bibfield{author}{\bibinfo{person}{Dong Su}, \bibinfo{person}{Huan Zhang},
  \bibinfo{person}{Hongge Chen}, \bibinfo{person}{Jinfeng Yi},
  \bibinfo{person}{Pin-Yu Chen}, {and} \bibinfo{person}{Yupeng Gao}.}
  \bibinfo{year}{2018}\natexlab{}.
\newblock \showarticletitle{Is Robustness the Cost of Accuracy?--A
  Comprehensive Study on the Robustness of 18 Deep Image Classification
  Models}. In \bibinfo{booktitle}{\emph{Proceedings of the European Conference
  on Computer Vision (ECCV)}}. \bibinfo{pages}{631--648}.
\newblock


\bibitem[Szegedy et~al\mbox{.}(2014)]%
        {szegedy2014intriguing}
\bibfield{author}{\bibinfo{person}{Christian Szegedy},
  \bibinfo{person}{Wojciech Zaremba}, \bibinfo{person}{Ilya Sutskever},
  \bibinfo{person}{Joan Bruna}, \bibinfo{person}{Dumitru Erhan},
  \bibinfo{person}{Ian~J. Goodfellow}, {and} \bibinfo{person}{Rob Fergus}.}
  \bibinfo{year}{2014}\natexlab{}.
\newblock \showarticletitle{Intriguing properties of neural networks}. In
  \bibinfo{booktitle}{\emph{Proceedings of the 2nd International Conference on
  Learning Representations}}, \bibfield{editor}{\bibinfo{person}{Yoshua Bengio}
  {and} \bibinfo{person}{Yann LeCun}} (Eds.).
\newblock


\bibitem[Tjeng et~al\mbox{.}(2019)]%
        {tjeng2018evaluating}
\bibfield{author}{\bibinfo{person}{Vincent Tjeng}, \bibinfo{person}{Kai~Y
  Xiao}, {and} \bibinfo{person}{Russ Tedrake}.}
  \bibinfo{year}{2019}\natexlab{}.
\newblock \showarticletitle{Evaluating Robustness of Neural Networks with Mixed
  Integer Programming}. In \bibinfo{booktitle}{\emph{International Conference
  on Learning Representations}}.
\newblock


\bibitem[Tram{\`{e}}r et~al\mbox{.}(2020)]%
        {tramer2020adaptive}
\bibfield{author}{\bibinfo{person}{Florian Tram{\`{e}}r},
  \bibinfo{person}{Nicholas Carlini}, \bibinfo{person}{Wieland Brendel}, {and}
  \bibinfo{person}{Aleksander Madry}.} \bibinfo{year}{2020}\natexlab{}.
\newblock \showarticletitle{On Adaptive Attacks to Adversarial Example
  Defenses}. In \bibinfo{booktitle}{\emph{Advances in Neural Information
  Processing Systems 33}}.
\newblock


\bibitem[Tsipras et~al\mbox{.}(2019)]%
        {tsipras2018robustness}
\bibfield{author}{\bibinfo{person}{Dimitris Tsipras}, \bibinfo{person}{Shibani
  Santurkar}, \bibinfo{person}{Logan Engstrom}, \bibinfo{person}{Alexander
  Turner}, {and} \bibinfo{person}{Aleksander Madry}.}
  \bibinfo{year}{2019}\natexlab{}.
\newblock \showarticletitle{Robustness May Be at Odds with Accuracy}. In
  \bibinfo{booktitle}{\emph{International Conference on Learning
  Representations}}.
\newblock


\bibitem[Turner et~al\mbox{.}(2019)]%
        {turner2019label}
\bibfield{author}{\bibinfo{person}{Alexander Turner}, \bibinfo{person}{Dimitris
  Tsipras}, {and} \bibinfo{person}{Aleksander Madry}.}
  \bibinfo{year}{2019}\natexlab{}.
\newblock \bibinfo{title}{Label-consistent backdoor attacks}.
\newblock
\newblock
\showeprint[arxiv]{1912.02771}


\bibitem[Vinayakumar et~al\mbox{.}(2019)]%
        {vinayakumar2019robust}
\bibfield{author}{\bibinfo{person}{R Vinayakumar}, \bibinfo{person}{Mamoun
  Alazab}, \bibinfo{person}{KP Soman}, \bibinfo{person}{Prabaharan
  Poornachandran}, {and} \bibinfo{person}{Sitalakshmi Venkatraman}.}
  \bibinfo{year}{2019}\natexlab{}.
\newblock \showarticletitle{Robust intelligent malware detection using deep
  learning}.
\newblock \bibinfo{journal}{\emph{IEEE Access}}  \bibinfo{volume}{7}
  (\bibinfo{year}{2019}), \bibinfo{pages}{46717--46738}.
\newblock


\bibitem[Wang et~al\mbox{.}(2019)]%
        {wang2019neural}
\bibfield{author}{\bibinfo{person}{Bolun Wang}, \bibinfo{person}{Yuanshun Yao},
  \bibinfo{person}{Shawn Shan}, \bibinfo{person}{Huiying Li},
  \bibinfo{person}{Bimal Viswanath}, \bibinfo{person}{Haitao Zheng}, {and}
  \bibinfo{person}{Ben~Y Zhao}.} \bibinfo{year}{2019}\natexlab{}.
\newblock \showarticletitle{Neural cleanse: Identifying and mitigating backdoor
  attacks in neural networks}. In \bibinfo{booktitle}{\emph{2019 IEEE Symposium
  on Security and Privacy (SP)}}. IEEE, \bibinfo{pages}{707--723}.
\newblock


\bibitem[Wang et~al\mbox{.}(2020)]%
        {wang2020practical}
\bibfield{author}{\bibinfo{person}{Ren Wang}, \bibinfo{person}{Gaoyuan Zhang},
  \bibinfo{person}{Sijia Liu}, \bibinfo{person}{Pin-Yu Chen},
  \bibinfo{person}{Jinjun Xiong}, {and} \bibinfo{person}{Meng Wang}.}
  \bibinfo{year}{2020}\natexlab{}.
\newblock \showarticletitle{Practical detection of trojan neural networks:
  Data-limited and data-free cases}. In \bibinfo{booktitle}{\emph{Computer
  Vision--ECCV 2020: 16th European Conference, Glasgow, UK, August 23--28,
  2020, Proceedings, Part XXIII 16}}. Springer, \bibinfo{pages}{222--238}.
\newblock


\bibitem[Xu et~al\mbox{.}(2020)]%
        {xu2020automatic}
\bibfield{author}{\bibinfo{person}{Kaidi Xu}, \bibinfo{person}{Zhouxing Shi},
  \bibinfo{person}{Huan Zhang}, \bibinfo{person}{Yihan Wang},
  \bibinfo{person}{Kai-Wei Chang}, \bibinfo{person}{Minlie Huang},
  \bibinfo{person}{Bhavya Kailkhura}, \bibinfo{person}{Xue Lin}, {and}
  \bibinfo{person}{Cho-Jui Hsieh}.} \bibinfo{year}{2020}\natexlab{}.
\newblock \showarticletitle{Automatic perturbation analysis for scalable
  certified robustness and beyond}.
\newblock \bibinfo{journal}{\emph{Advances in Neural Information Processing
  Systems}}  \bibinfo{volume}{33} (\bibinfo{year}{2020}).
\newblock


\bibitem[Zhang et~al\mbox{.}(2020)]%
        {zhang2020towards}
\bibfield{author}{\bibinfo{person}{Huan Zhang}, \bibinfo{person}{Hongge Chen},
  \bibinfo{person}{Chaowei Xiao}, \bibinfo{person}{Sven Gowal},
  \bibinfo{person}{Robert Stanforth}, \bibinfo{person}{Bo Li},
  \bibinfo{person}{Duane Boning}, {and} \bibinfo{person}{Cho-Jui Hsieh}.}
  \bibinfo{year}{2020}\natexlab{}.
\newblock \showarticletitle{Towards Stable and Efficient Training of Verifiably
  Robust Neural Networks}. In \bibinfo{booktitle}{\emph{International
  Conference on Learning Representations}}.
\newblock


\bibitem[Zhang et~al\mbox{.}(2018)]%
        {zhang2018efficient}
\bibfield{author}{\bibinfo{person}{Huan Zhang}, \bibinfo{person}{Tsui-Wei
  Weng}, \bibinfo{person}{Pin-Yu Chen}, \bibinfo{person}{Cho-Jui Hsieh}, {and}
  \bibinfo{person}{Luca Daniel}.} \bibinfo{year}{2018}\natexlab{}.
\newblock \showarticletitle{Efficient Neural Network Robustness Certification
  with General Activation Functions}.
\newblock \bibinfo{journal}{\emph{Advances in Neural Information Processing
  Systems}}  \bibinfo{volume}{31} (\bibinfo{year}{2018}).
\newblock


\bibitem[Zhong et~al\mbox{.}(2020)]%
        {zhong2020backdoor}
\bibfield{author}{\bibinfo{person}{Haoti Zhong}, \bibinfo{person}{Cong Liao},
  \bibinfo{person}{Anna~Cinzia Squicciarini}, \bibinfo{person}{Sencun Zhu},
  {and} \bibinfo{person}{David Miller}.} \bibinfo{year}{2020}\natexlab{}.
\newblock \showarticletitle{Backdoor embedding in convolutional neural network
  models via invisible perturbation}. In \bibinfo{booktitle}{\emph{Proceedings
  of the Tenth ACM Conference on Data and Application Security and Privacy}}.
  \bibinfo{pages}{97--108}.
\newblock


\end{thebibliography}

\appendix

\section{Poison Ratio}
\label{sec:experiments_poison_rate}

We show the influence of different poison ratios on the success of our indirect attack by running the same experiment introduced in \cref{sec:experiments_attacks} with $\epsilon=0.02$ and natural training, but for different poison ratios.
The fewer poisoned samples we add to the training data, the less likely the attack will be detected.

\begin{figure}
    \centering
    \includegraphics[width=\columnwidth]{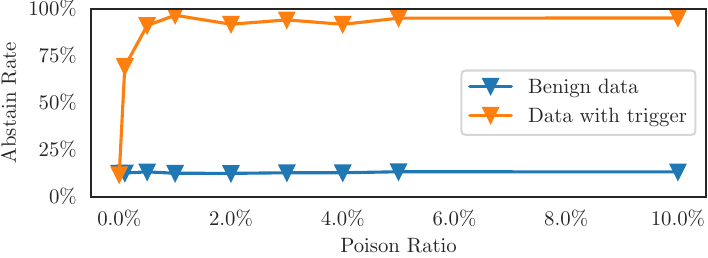}
    \vspace{-2em}
    \caption{Abstain rate on benign data and data with a trigger for different poison ratios.
    Fully connected networks trained on MNIST with our indirect attack.}
    \vspace{-1em}
    \label{fig:poison_ratio}
\end{figure}

\Cref{fig:poison_ratio} shows the abstain rate on benign data and data with a trigger for different poison ratios.
Adding just $0.5\%$ poisoned samples is already sufficient to increase the abstain rate from originally $11.7\%$ to $91.1\%$ on the trigger distribution.
Increasing the poisoning ratio to $1\%$ further increases the abstain rate to $96.5\%$, which remains in the same range for larger ratios. These results show that the attack is already highly effective for a small number of poisoned samples, hiding it well from the victim.

\section{Defenses}
\label{sec:experiments_defenses}
\balance
Given the high success rate of our attacks and their impact on deep learning systems, it is prudent to develop defenses against these threats.
Previous work on training-time attacks and defenses target \textit{accuracy} and not \textit{certification}.
Hence, it is unclear whether these methods can be adapted or if we require new defenses for availability attacks.
While this work primarily focuses on showing the vulnerability of certifiers to training-time attacks, we take the first step toward defenses.
We analyze the effectiveness of three defenses against traditional attacks in our novel setting: fine-pruning~\cite{liu2018fine}, neural cleanse~\cite{wang2019neural}, and trojan network detection~\cite{wang2020practical}.

\bparagraph{Fine-pruning}
Fine-pruning consists of two steps:
On a set of benign data, dormant neurons are pruned from the model to remove the trigger-related neurons.
The pruned model is then fine-tuned on the same subset.

\begin{table}
\centering
\begin{tabular*}{\columnwidth}{@{}l @{\extracolsep{\fill}} llll@{}}
    \toprule
    Neurons Pruned & 0\% & 25\% & 50\% & 75\% \\
    \midrule
    Benign Data \\
    \quad \textcolor{acc}{Accuracy} & \textcolor{acc}{\res{98.4}} & \textcolor{acc}{\res{98.5}}  & \textcolor{acc}{\res{98.3}}  & \textcolor{acc}{\res{97.8}} \\
    \quad Abstain Rate & \res{13.4} & \res{12.5}  & \res{11.0}  & \res{9.2} \\
    Data with Trigger \\
    \quad \textcolor{acc}{Accuracy} & \textcolor{acc}{\res{29.3}} & \textcolor{acc}{\res{25.2}}  & \textcolor{acc}{\res{36.4}}  & \textcolor{acc}{\res{63.6}} \\
    \quad Abstain Rate & \res{84.4} & \res{63.7}  & \res{68.2}  & \res{62.0} \\
    \bottomrule
\end{tabular*}

\vspace{.3em}
\caption{Accuracy and abstain rate for natural training of a fully-connected model on MNIST with $\epsilon = 0.02$ and different percentages of pruned connections.}
\vspace{-2em}
\label{tab:defense}
\end{table}

\Cref{tab:defense} shows accuracy and abstain rate with $\epsilon=0.02$ for an MNIST classifier trained with natural training and our indirect attack for different percentages of pruned connections.
With an increasing percentage of pruned neurons, the defense is able to recover some accuracy and abstain rate on data with a trigger, reaching 63.6\% accuracy and 62.0\% abstain rate when 96 (75\%) neurons have been pruned.
This is, however, still significantly below the target accuracy of 97.8\% and below the target abstain rate of 9.2\%.

\bparagraph{Neural Cleanse}
Neural cleanse~\cite{wang2019neural} is a more powerful defense against backdoor attacks, which can detect, identify, and remove backdoors.
It works in multiple stages, where the first stage detects the trigger by finding the minimal perturbation which misclassifies samples to a target label.
The trigger is detected by finding outliers in the magnitude of perturbation required for different labels.

Running this detection step on a network trained with our availability attack yields no outliers, and therefore the detection fails.
Since all consecutive steps rely on finding the perturbation pattern, the mitigation step cannot be applied.

This result makes sense since our attack does not cause misclassification to a particular target label, and therefore we would not expect decision boundaries to one target class near all others.

\bparagraph{Trojan Network Detection}
The third, recently published defense we evaluate is trojan network detection (TND)~\cite{wang2020practical}.
It detects backdoors in neural networks using feature inversion, exploiting the strong neuron activations at certain coordinates of trojan networks.
TND then compares the logit activations of these reverse-engineered inputs to those of benign inputs to flag the model as malicious.

This defense also fails to detect our attack.
When comparing the changes in logit activation, we observe that the change is similar in magnitude across all logits.
This can again be explained by the fact that we target the abstain rate of all classes instead of a single class.

\bparagraph{Discussion}
The results on all three defenses show that our availability attacks on certifiers differ significantly from traditional training-time attacks.
The key difference is that our attacks do not target misclassification and therefore require new approaches for effective defenses.
None of the evaluated defenses, designed to prevent misclassification, were able to detect or mitigate our novel attack, highlighting the need for customized solutions.

\end{document}